# Analytical Formulation of Autonomous Vehicle Freeway Merging Control with State-Dependent Discharge Rates

Qing Tang, Xianbiao Hu

*Abstract*—**The core of the freeway merging control problem lies in the dynamic queue propagation and dissipation corresponding to the behavior and state of merging vehicles. Traditionally, queuing has been modeled through demand-supply interactions, with time-varied demand and fixed capacity as supply. However, field observations indicate that flow rates decrease at freeway merging areas during congestion, primarily due to the impact of intersecting traffic—a factor overlooked in the fundamental diagram. This manuscript introduces an analytical approach to characterize and control the dynamic multi-stage merging process of autonomous vehicles, prioritizing traffic flow efficiency and safety. For the first time, the effective discharge rate at the merging point, reduced by the multi-stage dynamic merging process, is analytically derived using an explicit closed-form formulation. Leveraging this expression, we derive dynamic traffic flow performance measurements, including queue length and traffic delay, which serve as the first objective. Additionally, a crash risk function is established to quantitatively assess potential collisions during the merging process, serving as the second objective. Finally, we formulate this problem as a dynamic programming problem to jointly minimize delay and crash risk, with the merging location and speed as decision variables. Given the terminal state's availability, we formulate the ramp vehicle merging task as a recursive optimization problem, employing backward induction to find the minimum-cost solution. Numerical experiments are conducted to validate the derived effective discharge rate using the NGSIM dataset. The results indicate that the proposed model outperforms two benchmark algorithms, leading to a more efficient and safer merging process.**

*Index Terms*— **Autonomous Vehicle, Dynamic Queuing, Dynamic Programming, Effective Discharge Rate, Freeway Merging**

## I. Introduction

THE complex and dynamic interaction among vehicles during freeway merging contribute significantly to traffic congestion, accidents, and flow oscillation. Numerous control algorithms for autonomous vehicles (AVs) have been developed in the literature to assist with their merging maneuvers. Earlier studies focused on generating acceptable gaps in automated highway systems (AHS) to facilitate safe and smooth merging of ramp vehicles into mainline traffic [1-3]. Recent advancements in communication technologies have enabled more sophisticated merging algorithms leveraging inter-vehicle communications [4-6]. Key objectives of the freeway merging control, as identified in the literature, include increasing throughput, reducing delays, decreasing crash risk, lowering fuel consumption, minimizing engine effort, and reducing passenger discomfort. Control strategies encompass adjusting mainline speeds, controlling merging vehicle speeds, providing lane-changing advisories, promoting cooperative merging, and establishing platooning formations, among other strategies [3-5, 7, 8].

The freeway merging control problem centers on dynamic queue propagation and dissipation corresponding to the actions and states of the merging vehicle. Following classic traffic fundamentals, traffic dynamics have long been modeled as a result of demand-supply interactions, with time-varied demand and fixed capacity. The queuing profile is formulated as the integral of the difference between the time-varied demand or arrival rate and constant capacity. Mathematically, this is formulated as $Q(t) = \int_{t_0}^{t} (\lambda(\tau) - \mu)\, d\tau$, in which $Q(t)$ is the time-dependent queue, $\lambda(\tau)$ is the time-dependent arrival rate, $t_0$ is the start of the congestion period, and $\mu$ represents the constant capacity, or the maximum achievable discharge rate. This constant-capacity-based formulation approach is widely used in the morning commute problem [9] and associated research [10-12]. It significantly simplifies the derivation of closed-form queue profiles, vehicle delay, and the scheduling of optimal departure time for equilibrium analysis. Newell [13] further demonstrated consistency between simplified car-following models and triangular fundamental diagrams, enabling the calculation of maximum discharge rates and backward wave speed. This formulation underpins numerous vehicle control algorithms and queueing models, including works by Ahn, Cassidy [14], Chen, Laval [15], Wei, Avcı [16], and others.

Field observations, however, present a different picture. Using the I-80 NGSIM dataset in California as an example, **Fig. 1**-Left shows volume and speed profiles from loop detector #6, located in a freeway ramp merging area. The green dots indicate the discretized flow rate detection, while the blue curve depicts the continuous and smoothed flow profile. The red curve represents the speed profile, highlighting severe congestion between 8 am and 10 am, as evidenced by a substantial speed reduction. During this period, the loop detector register fewer vehicles (refer to the black rectangular box in **Fig. 1**-Left). Traffic recovery begins around

Qing Tang is with the Department of Civil and Environmental Engineering, Old Dominion University, Norfolk, 23529 USA (e-mail: q1tang@odu.edu)
Xianbiao Hu is with the Department of Civil and Environmental Engineering, University Park, PA 16802 USA (e-mail: xbhu@psu.edu)

10 am, with increasing speeds and flow rates, followed by a drop after queued vehicles are discharged. This reduction in capacity during congestion aligns with prior findings [17]. In merging areas, lane-changers from on-ramps cause main-road vehicles ("followers") to slow down to maintain safe distances. Once the merging vehicle has been fully accelerated or left the lane, the follower accelerates. However, the new time headway $h_{aft}$ is significantly greater than the initial headway $h_{bef}$, and may never return to $h_{bef}$ (see **Fig. 1**-Right). Consequently, the effective discharge rate during congestion decreases due to the impact of the lane-changer in this merging area.

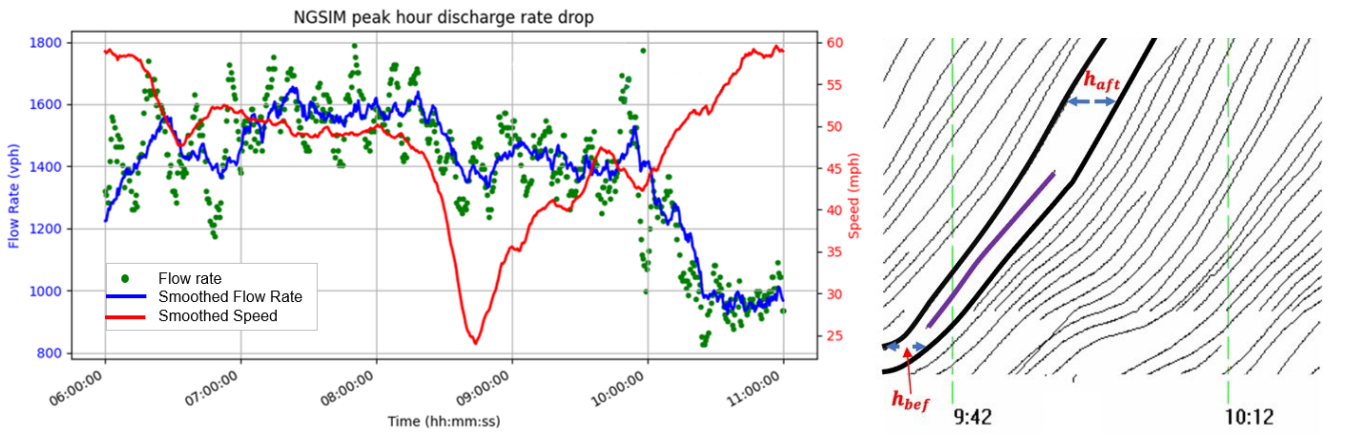

**Fig. 1.** Evidence of capacity drop in ramp merging area using NGSIM dataset. Left: speed and flow profile; Right: time-space diagram for headway analysis.

However, the fundamental diagram does not account for interactions arising from the flow of intersecting links, as it traditionally represents the relationship within a single link. Specifically, it relates changes in outflow rates to traffic density on that link, with density and flow represented on the X and Y axes, respectively. Conversely, **Fig. 1**-Right illustrates that vehicles from intersecting links influence the freeway mainline's discharge rate. To capture this, the fundamental diagram could be expanded into three dimensions: freeway density (X-axis), ramp traffic state (Y-axis), and freeway flow rate (Z-axis). **Fig. 1**-Right further suggests that the number of lane-changing vehicles affects the frequency and size of larger headway gaps, which are influenced by their speeds. Consequently, the traditional queue length equation $Q(t)$ mentioned earlier, which assums a constant $\mu$, should be modified for merging sections to $Q(t) = \int_{t_0}^{t}(\lambda(\tau) - \mu')d\tau$, where $\mu'$ depends on ramp vehicle ratio $\rho^r$ and their speed $v_r$.

An analytical closed-form formulation of $\mu'$, which characterizes the complex interactions at ramp merging points, is essential for effective control and optimization. Without such a formulation, it is impossible to mathematically link the merging vehicles' states and actions with queue propagation and dissipation process, and further derive evaluation metrics like delay and fuel consumption. Previous studies have modeled the capacity drop due to the dynamic interactions among vehicles. For example, Leclercq, Laval [18] modeled the capacity drop related to the merging process and vehicle performance heterogeneity. Menendez and Daganzo [19] used simulations to study bottleneck capacity reductions arising from high occupancy vehicle lanes, and Chen and Ahn [20] investigated how spatially distributed lane changes impact capacity drop at an extended merge, diverge, and weave bottlenecks. Other related works can be found in [21-26], and others. To the best of our knowledge, our previous work is the only research to analytically derive a closed-form $\mu'$ for specific cases: the bus-only lane conversion problem [17], where passenger vehicles merge from a side-street entrance onto the main road, and a roundabout scenario [27], where merging vehicles interact with those already in the roundabout.

This manuscript introduces an analytical approach to address this research gap by first characterizing and subsequently controlling the dynamic multi-stage merging process of an autonomous vehicle, prioritizing traffic flow efficiency and merging safety. We begin by outlining the multi-stage dynamic merging process and the underlying traffic dynamics, using Newell's simplified car-following model and time-space diagram. Different merging strategies are discussed, highlighting the trade-off between traffic flow efficiency and safety. The dynamic queuing process when a vehicle merges into the mainline and results in a capacity drop is analyzed, along with the metrics used for evaluation. Proceeding with this theoretical analysis, we develop mathematical models to formulate and control an AV's merging decisions. Using a closed-form expression with a capacity discount factor, we model the effective discharge rate at merging locations. Adopting the "Barrel Effect" concept, we identify the reference location with the lowest effective discharge rate. The merging vehicle's state and transitions are summarized in a state transition diagram. We then formulate a dynamic programming problem to optimize the merging decisions of the AVs, effectively integrating the dual objectives of traffic flow efficiency and safety. Given the terminal state's availability, we formulate the ramp vehicle merging task as a recursive optimization problem and solve it using backward induction to find the minimum-cost solution.

This paper is organized as follows: Section II reviews the literature and highlights research gaps. Section III covers the preliminaries, including the dynamic multi-stage merging process, the efficiency-safety trade-off, the dynamic queuing process, and evaluation metrics. Section IV introduces the core model, deriving a closed-form effective discharge rate at merging locations, defining vehicle states and decision variables, and formulating a recursive optimization problem to minimize delays and crash risks. Section V presents the model implementation and results, while Section VI concludes with final remarks.

## II. Literature Review

The literature offers extensive research on freeway ramp merging. Early studies focused on AHS and techniques for safer and more efficient merging. For instance, Ran, Leight [1] and Lu, Tan [2] developed gap-creation strategies using microscopic simulation models and merging control algorithms to facilitate smooth and safe merging, a method also adopted in later studies [3, 28]. With advances in information and communication technology, researchers began integrating inter-vehicle communication into merging control algorithms. Wang, Wenjuan [29] introduced a cooperative control algorithm that shares position and driving data to enable collision-free ramp merging. Awal, Kulik [30] optimized merging order between vehicle streams for improved efficiency. The advent of AVs and connected and autonomous vehicle (CAV) technologies has further advanced merging optimization. AVs can estimate merging intentions, predict strategies, and make decisions using cost functions that balance comfort, safety, and fuel consumption [31]. Cao, Mukai [32] proposed trajectory control for mainline AVs to ensure smoother merging and reduce congestion. Cooperative merging approaches, such as optimal control problems for CAV coordination, aim to minimize engine effort, enhance passenger comfort, and reduce travel time [5]. Marinescu, Čurn [33] further emphasized collaboration between mainline and on-ramp vehicles to improve throughput and reduce delays.

Various control strategies have been proposed to address merging and traffic flow optimization, including adjusting mainline speeds, promoting cooperative merging, and forming platoons. Milanés, Godoy [3] developed an automated system that adjusts mainline vehicle speeds to enable smooth merging and reduce congestion. Ntousakis, Nikolos [5] proposed a cooperative merging method that minimizes engine effort and passenger discomfort by optimizing CAV trajectories. Mu, Du [8] introduced a rolling horizon approach for safe and efficient merging of CAV and mixed platoons at mainline-ramp intersections. Roadside units were tested in another study [4] to compute and transmit optimized trajectories, leading to a proactive merging control algorithm that maximizes travel speed. Additionally, Park, Bhamidipati [7] investigated a lane-changing advisory system encouraging early lane changes near ramps to improve speed and fuel efficiency. These strategies often aim to optimize metrics like traffic efficiency [4-7, 32-34] or reduce crash risks [1-3, 28-30, 35].

To evaluate the performance of merging control algorithms, metrics such as vehicle delay, travel time, and fuel consumption are commonly used. Accurate calculation of these metrics requires capturing the dynamic queue propagation and dissipation associated with the actions and states of merging vehicles. Using traffic flow principles, queuing profiles are typically derived as the integral of the difference between time-varying demand (or arrival rate) and fixed capacity [10-12, 14-16]. The phenomenon of capacity drop at freeway merging locations has been widely studied. For instance, Leclercq et al. [18] extended the Newell-Daganzo model to account for capacity drops caused by merging processes and vehicle heterogeneity but required solving predefined equations for specific parameters, as no explicit closed-form solution was provided. Other studies, such as Laval and Daganzo [36], analyzed how lane-changing vehicles create voids in traffic streams that reduce flow, assuming the probability of lane changes depends on positive speed differences but without considering on- and off-ramp traffic. Menendez and Daganzo [19] used simulations to study capacity reductions due to high-occupancy vehicle lanes, while Chen and Ahn [20] developed analytical and simulation models to assess capacity drops caused by distributed lane changes at merge, diverge, and weave bottlenecks. Although many studies have explored capacity drops [21, 22, 26, 37-39], most models either lack explicit analytical forms or rely heavily on simulations, limiting their ability to offer a clear analytical understanding of underlying traffic behaviors.

To summarize, despite its importance in characterizing traffic flow dynamics, a closed-form formulation for freeway merging capacity drop remains unavailable. This absence makes it impossible to analytically derive performance metrics related to the actions and states of merging vehicles. To our knowledge, our prior work is the only research to analytically formulate closed-form effective discharge rates for bus-only lane conversion [17] and roundabout merging scenarios [27]. Below, we describe the establishment of analytical models to fill this research gap.

## III. Preliminaries and Theoretical

This section presents the preliminaries and theoretical analysis to support the core models introduced in Section IV. We decompose an AV's decision-making at ramp merging point into a dynamic multi-stage merging process and analyze the inherent traffic dynamics. Based on this analysis, we further investigate the trade-offs between traffic flow efficiency and safety in merging decision, define merging vehicle states and their dynamic transitions, and introduce evaluation metrics. For clarity, a summary of the mathematical notations used is provided in Appendix A.

### *A. Multi-stage Merging Process with Repeating Episodes in a Time-space Diagram*

Our analysis focuses on a typical freeway segment with an on-ramp and an auxiliary lane, accommodating mixed traffic with AVs and human-driven vehicles (HDVs). Fig. 2-Left illustrates a physical representation of the merging scenario, where ramp intersects the freeway at the beginning of the auxiliary lane (point $\mathcal{O}$), which ends at point $\mathcal{P}$. Vehicles entering the auxiliary lane seek acceptable gaps to merge into the mainline. Due to the typical disparity between the speed limit on the ramp and the cruising speed on the mainline, merging vehicles often accelerate, with the merging point ($\mathcal{M}$) always located within segment $\mathcal{OP}$.

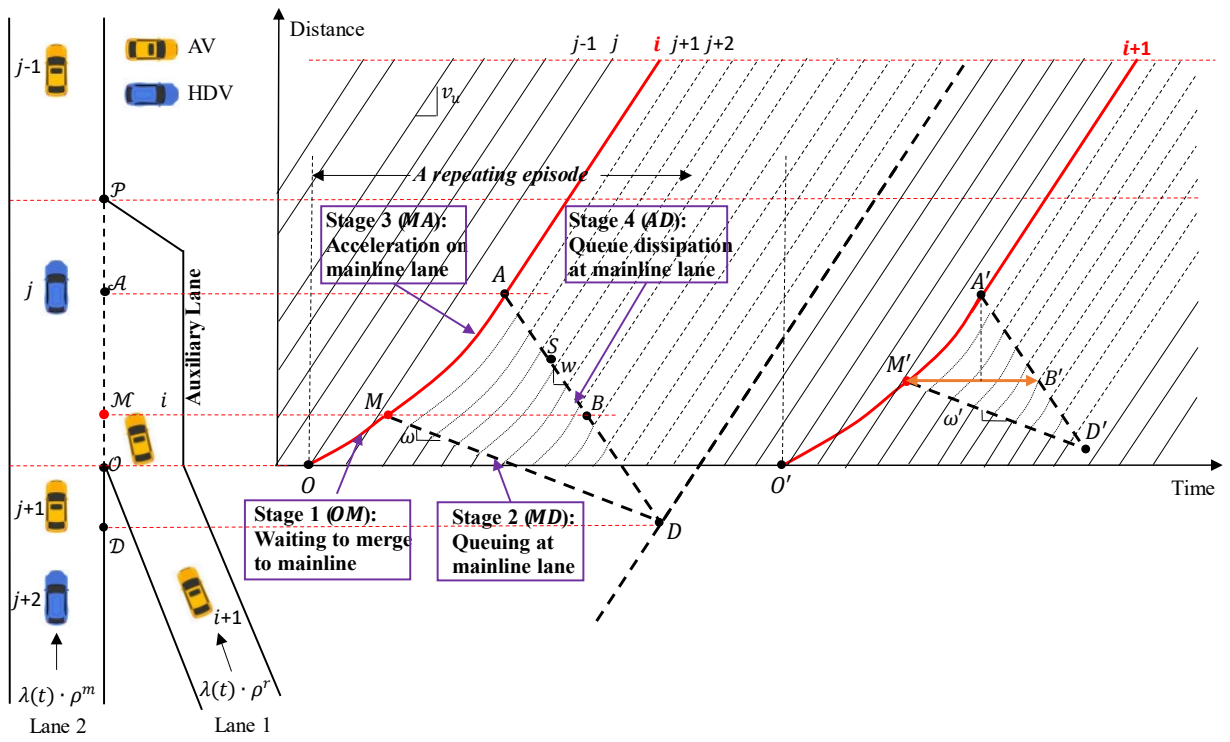

**Fig. 2**. Illustrations of the multi-stage vehicle merging process. Left: physical movement; Right: time-space diagram.

We focus on merging control of an AV from the on-ramp (highlighted in yellow in **Fig. 2**-Left). The freeway traffic consist of a mix of AVs and HDVs (marked in blue), exhibiting differences in characteristics such as perception-response times and car-following distances. The on-ramp traffic is also mixed with AVs and HDVs. However, HDVs on the ramp are considered uncontrollable, and their merging behavior is excluded from this study. AVs are assumed to make independent merging decisions, with no consideration of cooperative decision-making among them.

We adopt Newell's simplified car-following model [13], widely utilized to characterizing vehicle interactions at merging locations [17, 40, 41], to describe vehicle kinematic on freeways. The time-space diagram in **Fig. 2**-Right illustrates the merging process and vehicle interactions. Black curves represent freeway vehicles, whereas thick red curves (e.g., $OMA$ ) denote merging vehicles from on-ramps. Following the model, a following vehicle's trajectory is essentially the same as that of the leading vehicle, except for a temporal delay $\tau_n$ (reaction time) and a spatial delay $d_n$. For instance, if a leading vehicle $n-1$ transitions from speed $v_1$ to $v_2$ at time $t$, the following vehicle $n$ achieves the same speed, with its position at time $(t+\tau_n)$ given by $x_n(t+\tau_n)=x_{n-1}(t)-d_n$. Based on the time-space diagram, the time headway $h_n$ and space headway $s_n$ of vehicle $n$, before and after speed changes, transition from $h_{n,1}=\tau_n+d_n/v_1$ to $h_{n,2}=\tau_n+d_n/v_2$ and from $s_{n,1}=d_n+\tau_n\cdot v_1$ to $s_{n,2}=d_n+\tau_n\cdot v_2$, respectively. When the following vehicle is an AV, $\tau_n$ and $d_n$ are typically lower than those for HDVs.

We decompose the freeway merging process into multiple stages, as shown in Fig. 2-Right. An AV $i$ arrives at the entrance of the auxiliary lane at time $t_O$ with speed $v_O$. In stage 1, it accelerates on the auxiliary lane and seeks a merging opportunity (segment $OM$ in Fig. 2-Right). If an acceptable gap, encompassing a lead and a lag gap, emerges, this vehicle may merge into the freeway mainline at point $M$ with speed $v_M$, or wait for another opportunity. In stage 2, it continues to accelerate on the freeway mainline, and as its speed is slower than the freeway cruising speed $v_u$, a queuing wave is formed at point $M$ and propagates backward (segment $MD$). The vehicle accelerates until reaching $v_u$ at point $A$. Once reaching $v_u$, a new shockwave originates and propagates backward, dissipating the queue. This corresponds to segment $MA$, referred to as stage 3. The two shockwaves intersect at point $D$, where the merging vehicle no longer impacts traffic flow. If traffic becomes more saturated, point $D$ shifts downstream. In extreme cases, if the shockwaves do not intersect, this stage 4 is skipped. Locations $\mathcal{D},\mathcal{O},\mathcal{M},\mathcal{A},\mathcal{P}$ in **Fig. 2**-Left correspond to points $D,O,M,A,P$ in **Fig. 2**-Right.

A repeating episode is defined using the time-space diagram in **Fig. 2**-Right, with its duration determined accordingly. The multi-stage merging process repeats when the next merging vehicle, $i+1$, arrives at the entrance $O$ at time$t_{O'}$, with its trajectory shown in red. The time interval $[t_O,t_{O'}]$ signifies a complete episode, reflecting repeating patterns during the analysis period. This duration also represents the time headway of merging vehicles and is expressed as $h_r=1/\lambda(t)\cdot\rho^r$, where $\lambda(t)$ is the overall arrival rate and $\rho^r$ is the ratio of merging vehicles from the on-ramp.

### *B. Merging Strategies and Efficiency-safety Trade-off*

As illustrated in **Fig. 2**, a ramp vehicle can choose its merging point $M$ anywhere between points $\mathcal{O}$ and $\mathcal{P}$, influencing both traffic flow efficiency and driving safety. This subsection examines the safety requirements for lane changing during merging and discuss the trade-off of various merging strategies.

Considering the requirement for avoiding collision, the required time headway consists of a critical lag gap and a required lead gap, which are influenced by the driving speed $v(t)$, the vehicle's maximum deceleration $\alpha$, and the driver's response time $\tau$. If the following vehicle is an AV, $\tau$ is significantly reduced, being dictated by communication time and software responsiveness [42]. The critical lag gap for merging is given by $t^c_{lag}(t)=\tau+v_u/\alpha$, where $v_u$ is the freeway cruising speed. In terms of the required lead gap, since AVs do not need the response time $\tau$, it is given by $t^c_{lead}(t)=v(t)/\alpha$. Therefore, the total required time headway for merging can be computed as $h^c_M(t)=\tau+v_u/\alpha+v(t)/\alpha$. Next, we analyze the probability of finding an acceptable gap. In the classic traffic flow theory, vehicle arrivals are often assumed to follow a Poisson distribution, with vehicle arrival headways subjecting to a negative exponential distribution. Specifically, the headway probability density function is defined as $f(h;\lambda')=\begin{cases}\lambda' e^{-\lambda' h} & h\geq 0\\ 0 & h<0\end{cases}$,where $\lambda'=\lambda(t)\cdot\rho^m/3600$. Here $\rho^m$ is the ratio of cruising vehicles on the freeway mainline, with $\rho^m+\rho^r$ =1. As such, the probability of finding an acceptable gap for merging can be derived as $P_M(t)=e^{-\lambda'\cdot h^c_M(t)}$. By substituting in $\lambda'$ and $h_M(t)$, $P_M(t)$ can be obtained as in

$$P_M(t)=e^{-\left(\lambda(t)\cdot\frac{\rho^m}{3600}\right)\cdot\left(\tau+\frac{v_u}{\alpha}+\frac{v(t)}{\alpha}\right)} \tag{1}$$

The merging point decision significantly impacts freeway traffic flow. From an efficiency aspect, as illustrated in **Fig. 2**, the two shockwaves, $MD$ and $AD$, determine the number of vehicles affected by the merging vehicle and the resulting delays. Early merging at a lower speed can cause greater traffic disturbance, represented by a steeper $MD$, impacting more vehicles with longer delays. From a safety perspective, early merging results in a larger speed difference between the merging vehicle and freeway traffic, increasing the risk of sudden braking and accidents. Conversely, later merging at a higher speed minimizes traffic disruption. In an extreme case where points $A$ and $M$ overlap in **Fig. 2**, the ramp vehicle can merge at freeway cruising speed without creating shockwaves.

The selection of a merging point involves balancing the probability of finding an acceptable gap against its impact on traffic flow. Some studies adopted an "*early-merging strategy*" where vehicles would merge as soon as an acceptable gap appears Letter and Elefteriadou [4]. This approach increases probability of finding a gap due to the merging vehicle's slower speed, as indicated by $P_M(t)$ in (1), but it can disrupt traffic flow. Alternatively, a '*late--merging strategy'* has been explored [3, 5, 8], where vehicles bypasses earlier gaps on the auxiliary lane and merges only at the end of the lane (point $\mathcal{P}$). While this reduces traffic disturbance due to better vehicle acceleration, the required time headway ($h_M(t)$) increases with higher speeds, lowering the likelihood of finding a gap. If no acceptable gap is available at point $\mathcal{P}$, vehicles may be forced to merge, potentially causing crashes if freeway vehicles fail to slow down. These two strategies will be used for benchmarking purposes in Section V.

### *C. Dynamic Queuing with State-dependent Effective Discharge Rates and Evaluation Metrics*

In this subsection, we discuss the dynamic queuing process when a vehicle merges into the mainline and results in a capacity drop, as well as the metrics that we adopt to evaluate the outcomes. As discussed above, different merging strategies result in different queue propagation and dissipation processes. On one hand, this dynamic queuing process slows vehicles down, leading to traffic delays. On the other hand, the speed fluctuations in the following vehicles indicate crash risks. Continuous delay cost and crash risk functions, which reflect the resulting capacity drop and crash risks respectively, are presented below.

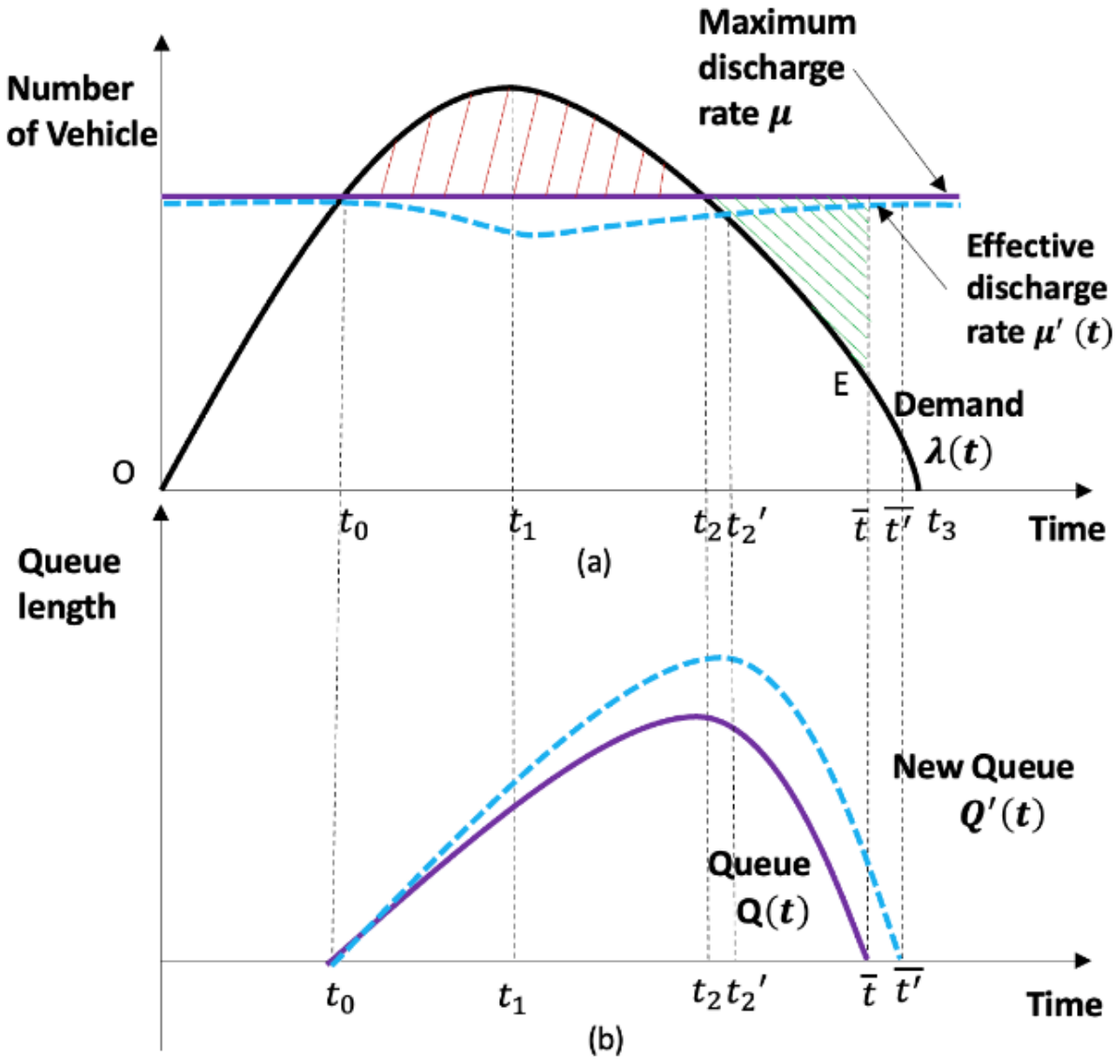

**Fig. 3.** Queue formation and dissipation process under a merging maneuver.

A fluid-based approximation method, as introduced by Newell [43], is utilized to calibrate the process of queue formation and dissipation. To begin, the traffic flow is characterized with the maximum discharge rate, $\mu$, which is typically regarded as a fixed value and corresponds to the highest flow rate in the triangle fundamental diagram. **Fig. 3** illustrates the demand-supply relationship and the resulting queueing profile. In **Fig. 3** (a), the demand (thick black solid curve), denoted as $\lambda(t)$, exceeds the maximum discharge rate $\mu$ (horizontal line in purple color) during peak hours between $t_0$ and $t_2$. The cumulative delay is represented by the area highlighted by red lines. Correspondingly, in **Fig. 3** (b), a queue (the purple curve $Q(t)$) begins to grow at time $t_0$, reaching the furthest upstream location at time $t_2$. Afterward, the queue begins to dissipate and is fully discharged by time $\bar{t}$. Using the maximum discharge rate as supply, the queue length at time $t$ is given by $Q(t)=\int_{t_0}^{t}(\lambda(\tau)-\mu)d\tau$, and the total delay is derived by $W(\mu)=\int_{t_0}^{t_3}Q(t)dt=\int_{t_0}^{\bar{t}}Q(t)dt$.

However, as previously discussed, the effective discharge rate, $\mu'$, is discounted after the ramp vehicles merge into the freeway mainline, resulting in a rate lower than $\mu$. Rather than adhering to the horizontal purple curve, the capacity discount is affected by various factors such as ramp flow rate, freeway cruising speed, and merging speed. Taking these variables into consideration, the effective discharge rate, $\mu'$, should be a time-dependent curve that remains consistently below the horizontal curve. This fluctuation is represented by the blue dashed curve in **Fig. 3** (a). Consequently, the queuing profile in **Fig. 3** (b) also requires an update. Instead of the purple curve $Q(t)$, the length of the queue increases due to the capacity discount and is represented by the blue dashed curve $Q'(t)$. Therefore, the longest queue occurs at time $t_2'$, and it is completely discharged at time $\bar{t}'$. The queue length at any time $t$ can be calculated as $Q'(t)=\int_{t_0}^{t}(\lambda(\tau)-\mu')d\tau$, and the total delay can be derived by

$$W(\mu')=\int_{0}^{t_3}Q'(t)dt=\int_{t_0}^{\bar{t}'}Q'(t)dt \tag{2}$$

In terms of safety considerations, the merging of ramp vehicles poses a risk of potential crashes, particularly when the merging gap is short. We adopt the deceleration rate to avoid a crash (DRAC), originally proposed by Cunto and Saccomanno [44], as an effective metric to estimate the driving risk of an individual vehicle. The formulation of DRAC is given in

$$DRAC=\frac{\left(v_j(t)-v_{j-1}(t)\right)^2}{x_{j-1}(t)-x_j(t)-L_{j-1}} \tag{3}$$

The index $j$ represents the following (response) vehicle, while $j-1$ denotes the lead (stimulus) vehicle. A vehicle's position is denoted by $x$ and $L$ represents the vehicle's length. The DRAC value quantifies crash risk, with higher values indicating greater likelihood. When a vehicle merges without ensuring an acceptable gap—such as a ramp vehicle reaching the end of an auxiliary lane without finding sufficient space—the inter-vehicle distance decreases. This reduction sharply increases the DRAC value, signaling risky behavior and imposing a significant penalty for unsafe driving.

Considering the entire multi-stage merging process, the crash risks associated with mering decision can be computed as the integral DRAC in

$$\mathcal{S}=\int_{t_O}^{t_{O'}}\left(\frac{\left(v_{j+1}(t)-v_i(t)\right)^2}{x_i(t)-x_{j+1}(t)-L_i}+\sum_{k=1}^{K}\frac{\left(v_{j+k+1}(t)-v_{j+k}(t)\right)^2}{x_{j+k}(t)-x_{j+k+1}(t)-L_{j+k}}\right)dt \tag{4}$$

Where $K$ is the number of vehicles affected by the merging behavior during one episode $OO'$, as shown in **Fig. 2**. The first component is the $DRAC$ between the merging vehicle $i$ and the following vehicle $j+1$, and the second component is the summation of $DRAC$ of all following affected vehicles.

## IV. Analytical Formulation and Control of Freeway Merging

This section introduces an analytical framework to formulate and then control an AV's decision during freeway merging. It begins by modeling the effective discharge rate at merging locations using a closed-form expression with a capacity discount factor. The merging vehicle's state and transitions are then defined, followed by the formulation a dynamic programming problem to optimize merging decisions, tightly integrating the dual objectives of traffic flow efficiency and safety.

### *A. Analytical derivation of effective discharge rate with a closed-form formulation*

We analyze a single episode, such as $OO'$ in **Fig. 2**, and compute the effective discharge rate as $\mu'=N_{OO'}/t_{OO'}$, where $N_{OO'}$ is the number of vehicles passing this merging area, and $t_{OO'}$ is the episode duration. As discussed earlier, the ramp vehicle experiences a multi-stage merging process until it is fully accelerated. **Fig. 4** illustrates how these four stages of the ramp vehicles lead to distinct freeway traffic states. These include: (1) a normal state ($TS_1$) where freeway traffic is unaffected by the ramp vehicle (blue area); (2) a no-vehicle-passing state ($TS_2$) where the ramp vehicle acts as a moving bottleneck, creating an empty space in the traffic flow (white area); (3) a queuing state ($TS_3$) where traffic jams and slows to follow the merging vehicle (orange area); and (4) a dissipation flow ($TS_4$) where queued vehicles dissipate at full capacity once the merging vehicle accelerates fully (green area). Once the jammed vehicles have fully dissipated, the flow returns to normal state ($TS_1$), indicated by the blue area. Subsequently, we can rewrite the equation as $\mu' = \sum N_{TS} / \sum t_{TS}$.

Next, we introduce the concept of the 'Barrel Effect' to examine the effective discharge rates along the freeway merging area. As the merging vehicle accelerates between points $\mathcal{M}$ and $\mathcal{A}$, it becomes a moving bottleneck that affects the freeway segment from physical point $\mathcal{D}$ to point $\mathcal{A}$ during the time period $[t_{M,} t_D]$ (as illustrated in **Fig. 4**). Due to the merging vehicle's changing speed, the degree of capacity drop varies based on the location. This is similar to the barrel effect, where the capacity of a barrel is determined by the height of its shortest stave rather than the longest. Likewise, a roadway's capacity is determined by its bottleneck, where the capacity is lowest. In other words, if we place a random point $Q$ along segment $\mathcal{DA}$, the capacity of the freeway merging area becomes $\mu' = \min\{\mu_Q, \forall Q \subseteq [\mathcal{D}, \mathcal{A}]\}$. Therefore, it is crucial to identify a reference location where the capacity drop is the most significant.

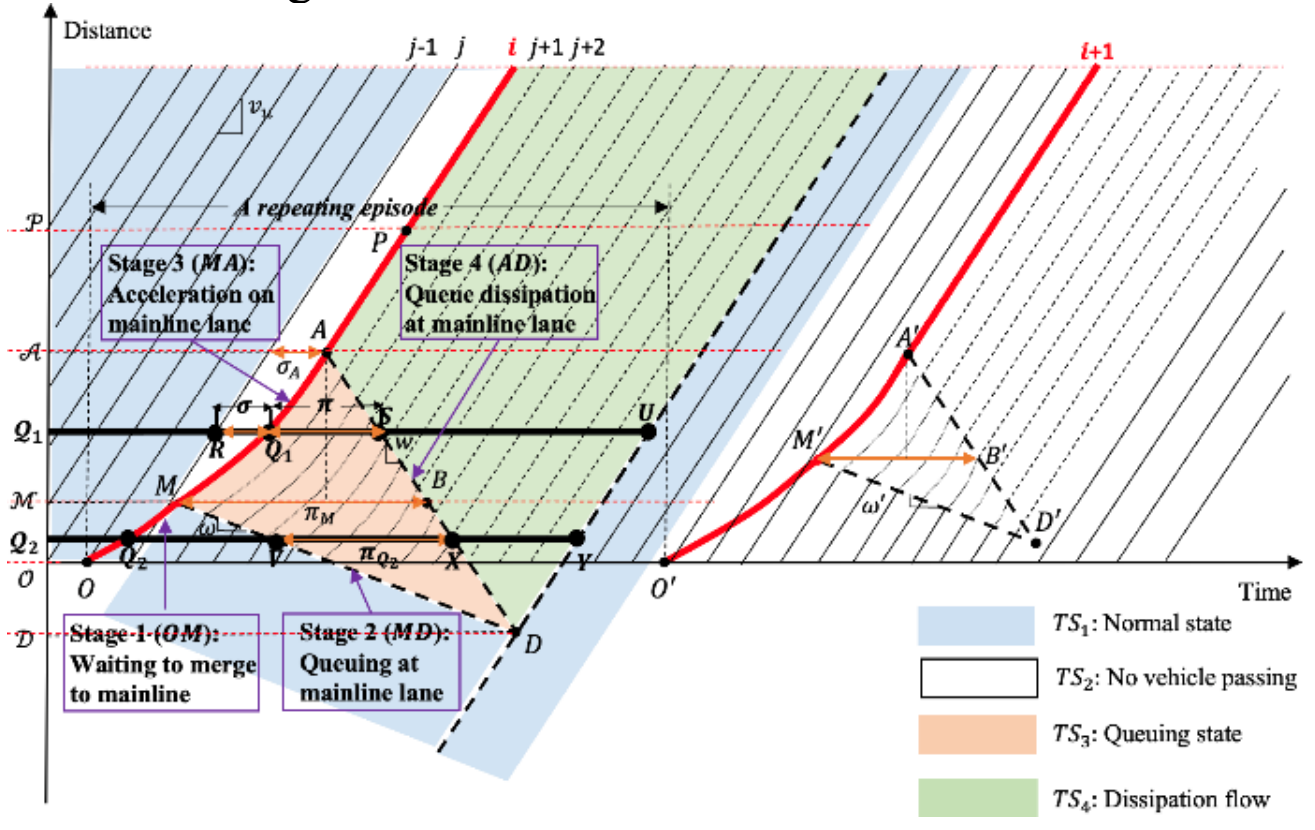

**Fig. 4.** Decomposition of freeway state with multistage merging process.

To identify a reference location where the effective discharge rate is the lowest, we need to examine different state transition chains when reference point $Q$ moves along the freeway segment. As an episode starts with a merging vehicle and ends with the arrival of a new merging vehicle, the chain always starts with a normal freeway state $TS_1$ and ends with state $TS_1$. A total of four state transition chains can be enumerated, and we use Fig. 5 to illustrate them. (1) Scenario 1: $TS_1 \rightarrow TS_2 \rightarrow TS_3 \rightarrow TS_4 \rightarrow TS_1$, illustrated by the red arrows in Fig. 5. This state transition chain is observed when reference point $Q$ is placed at subsegment $\mathcal{MA}$. (2) Scenario 2: $TS_1 \rightarrow TS_3 \rightarrow TS_4 \rightarrow TS_1$, illustrated by the blue arrows in Fig. 5. This state transition chain is observed when reference point $Q$ is placed at subsegment $\mathcal{DM}$. (3) Scenario 3: $TS_1 \rightarrow TS_1$, illustrated by the green arrow in **Fig. 5**. This state transition chain is observed when reference point $Q$ is placed at a location that is upstream of point $\mathcal{D}$. As illustrated in **Fig. 4**, at this upstream location, the freeway traffic is unimpacted by the merging vehicle from the on-ramp, as such, the traffic state remains at $TS_1$; and finally, (4) Scenario 4: $TS_1 \rightarrow TS_2 \rightarrow TS_4 \rightarrow TS_1$, illustrated by the yellow arrows in Fig. 5. This state transition chain is observed when reference point $Q$ is placed at a location that is downstream of point $\mathcal{A}$. As illustrated in **Fig. 4**, at this downstream location, the freeway traffic has stabilized again, as such, the traffic state $TS_3$ is skipped. In the following subsections, we will initially examine the more complex scenarios 1 and 2, before moving onto scenarios 3 and 4, and then summarizing the results.

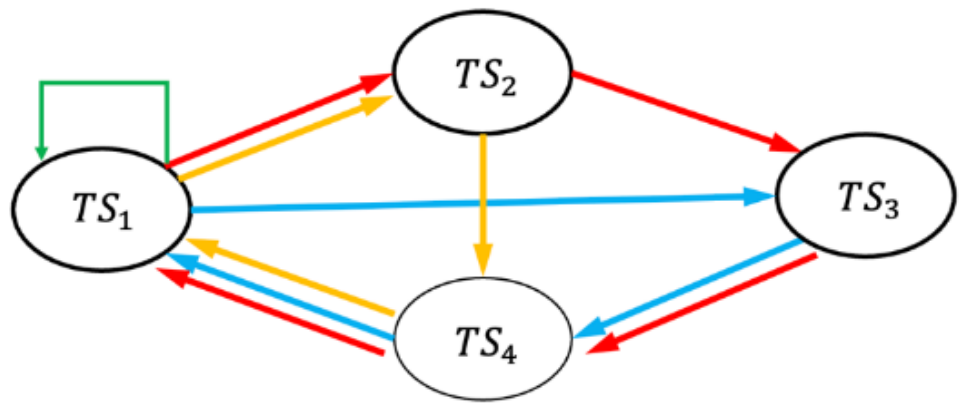

**Fig. 5.** State Transition chains for freeway traffic.

1) Scenario 1: reference point $Q$ located on subsegment $\mathcal{MA}$

To begin the analysis, we will focus on the first subsegment, $\mathcal{MA}$, and select a random point $Q_1$ on this subsegment as the reference point. The effective discharge rate at $Q_1$ during one episode, $[t_O, t_{O'}]$, can be broken down into five distinct periods: $[t_O, t_R]$, $[t_R, t_{Q_1}]$ which we denote as $\sigma_{Q_1}$, $[t_{Q_1}, t_S]$ which we denote as $\pi_{Q_1}$, $[t_S, t_U]$, and $[t_U, t_{O'}]$, each corresponding to a state in the transition chain $TS_1 \rightarrow TS_2 \rightarrow TS_3 \rightarrow TS_4 \rightarrow TS_1$. Point $R$ is observed when the last vehicle before the ramp vehicle passes point $Q_1$; point $S$ is observed when the discharging shockwave passes point $Q_1$; and point $U$ is observed when the last impacted vehicle passes point $Q_1$, as illustrated in **Fig. 4**. Subsequently, the effective discharge rate of one repeating episode, $\mu'$, can be expressed as $\mu' = (t_{OR} \cdot \mu_{OR} + \sigma_{Q_1} \cdot \mu_{RQ_1} + \pi_{Q_1} \cdot \mu_{Q_1S} + t_{SU} \cdot \mu_{SU} + t_{UO'} \cdot \mu_{UO'})/t_{OO'}$. The discharge rate and duration of each component will be analyzed next.

Emphasis is first placed on state $TS_3$, which corresponds to the area shaded in orange color in **Fig. 4**. After a reference point $Q_1$ is specified, segment $Q_1S$ becomes the period of interest and we denote its duration as $\pi_{Q_1}$. This interval signifies the moments when the merging vehicle impacts the traffic on the freeway. From time $t_M$ to $t_A$, the merging vehicle drives slower than vehicles on the freeway mainline, creating a moving bottleneck. During this period $\pi_{Q_1}$, vehicles can pass at a lower discharge rate. This period consists of two sub-periods: $Q_1A$, during which the merging vehicle keeps accelerating, and $AS$, during which the merging vehicle has been fully accelerated and drives at the cruising speed. The effective discharge rate for the $\pi_{Q_1}$ duration can be calculated using (5). The derivation process for each variable's value will be detailed subsequently.

$$\mu_{Q_1S} = ((t_A - t_{Q_1})\mu_{Q_1A} + (t_S - t_A)\mu_{AS})/\pi_{Q_1} \tag{5}$$

During the sub-period $Q_1A$, the merging vehicle is accelerating on the freeway mainline but has not reached the freeway cruising speed, $v_u$. As such, it becomes a moving bottleneck during the time $[t_{Q_1}, t_A]$, obstructing following vehicles. The flow rate passing through this moving bottleneck is zero, i.e., $\mu_{Q_1A} = 0$. Once the merging vehicle reaches point $A$, it has fully accelerated, and the flow is no longer impeded by the moving bottleneck. The previously queued vehicles then begin to discharge at a backward wave speed of $w$ and a density of $k_j$. The effective discharge rate can be derived as $\mu_{AS} = wk_j$. Therefore, $\mu_{Q_1S}$ in (5) simplifies to

$$\mu_{Q_1S} = wk_j(t_S - t_A)/\pi_{Q_1} \tag{6}$$

Next, the time difference between points $Q_1$, $A$ and $S$ need to be determined. As the speed of on-ramp is usually much lower than that of freeway, the ramp vehicle will accelerate as much as possible to match with the prevailing speed on freeway. We use $\alpha$ to denote the acceleration rate of the merging vehicle, which can be fine-tuned using field data. Since the merging vehicle reaches the cruising speed at point $A$, we have $t_A - t_{Q_1} = (v_u - v_{Q_1})/\alpha$.

We can then compute the time difference between $A$ and $S$. The spatial distance between point $A$ and $S$ is the same as the distance from point $Q_1$ to $A$, i.e., $x_{AS} = x_{Q_1A}$. From point $Q_1$ to $A$, we have $v_u = v_M + \alpha t_{Q_1A}$, and $x_{Q_1A} = v_M t_{Q_1A} + \alpha {t_{Q_1A}}^2/2$. Additionally, the interval between points $A$ and $S$ can be calculated using (7) referring to Appendix B.1.

$$t_S - t_A = x_{Q_1A}/w = (v_u^2 - v_{Q_1}^2)/2\alpha w \tag{7}$$

Therefore, the time interval, $\pi_{Q_1}$, can be derived as $\pi_{Q_1} = (v_u^2 - v_{Q_1}^2)/2\alpha w + (v_u - v_{Q_1})/\alpha$. With this, the effective discharge rate for traffic state $TS_3$, or $\mu_{Q_1S}$, as given in (6), can be rewritten in (8). Please refer to Appendix B.2 for details.

$$\mu_{Q_1S} = \frac{wk_j(v_u + v_{Q_1})}{v_u + v_{Q_1} + 2w} \tag{8}$$

Now that the effective discharge rate for state $TS_3$ has been derived, we will move onto state $TS_2$, another major period when the merging vehicle interacts with vehicles on the freeway. In **Fig. 4**, this corresponds to the segment between $R$ and $Q_1$ and we denote its duration as $\sigma_{Q_1}$. During period $\sigma_{Q_1}$(state $TS_2$), no vehicle can overtake the merging vehicle, which means the discharge rate $\mu_{RQ_1} = 0$. The speed at point $M$ is denoted as $v_M$ and the speed at point $Q_1$ is labeled as $v_{Q_1}$, such that $v_M \le v_{Q_1} \le v_\mu$. The time difference between point $R$ and $Q_1$, or $\sigma_{Q_1}$, can be derived as $\sigma_{Q_1} = (v_{Q_1} - v_M)/\alpha - (v_{Q_1}^2 - v_M^2)/2\alpha v_u$. The detailed derivation process is similar to Appendix B.1, and for the sake of simplicity, it is not presented here.

Proceeding to the remaining three time periods in the episode. It is observed that during $[t_O, t_R]$, the merging vehicle has no impact on traffic in the freeway mainline, so we have $\mu_{OR} = \mu$, although the actual flow rate might be lower than $\mu$ if the traffic is unsaturated. Next, for the segment $[t_S, t_U]$, since the ramp vehicle has been fully accelerated, the traffic flow is being discharged at full capacity, so we have $\mu_{SU} = \mu$. Finally, for segment $[t_U, t_{O'}]$, the impact of the merging vehicle has disappeared, and traffic on the freeway mainline returns to normal, so we have $\mu_{UO'} = \mu$. The duration of these three periods equals to the episode length $h_r$ minus the duration of the other two periods, i.e., $(t_R - t_O) + (t_U - t_S) + (t_{O'} - t_U) = \frac{1}{\lambda(t)\rho^r} - (\frac{v_{Q_1} - v_M}{\alpha} - \frac{v_{Q_1}^2 - v_M^2}{2\alpha v_u}) - (\frac{v_u - v_{Q_1}}{\alpha} + \frac{v_u^2 - v_{Q_1}^2}{2\alpha w})$.

Finally, by combining these five periods, the effective discharge rate of the freeway mainline over a complete episode can be computed by $\mu' = (\sigma_{Q_1}\mu_{RQ_1} + \pi_{Q_1}\mu_{Q_1S} + (t_{OR} + t_{SU} + t_{UO'})\mu)/t_{OO'}$. The final expression of $\mu'$ is simplified in (9), and please refer to Appendix B.3.

$$\mu' = \mu\left(1 - \lambda(t)\rho^r(t)\frac{(v_u - v_M)^2}{2\alpha v_u}\right) \tag{9}$$

A very interesting observation is that $v_{Q_1}$ cancels out during the derivation. This means that the selection of reference point $Q_1$ within segment $\mathcal{MA}$ does not impact the effective discharge rate, and that $\mu'$ is only impacted by vehicle arrival rate $\lambda(t)$, the ramp vehicle ratio $\rho^r(t)$, maximum acceleration rate $\alpha$, the speed of freeway vehicles $v_u$, and the speed of ramp vehicle merging into the freeway $v_M$. Another observation is that the first order derivative of $\mu'$ with regard to $v_M$ is always nonnegative, $\frac{\partial \mu'}{\partial v_M} = \mu\frac{\lambda(t)\rho^r(t)}{\alpha v_u}(v_u - v_M) \ge 0$. This indicates that when the reference point $Q_1$ is placed within segment $\mathcal{MA}$, $\mu'$ is a monotonically non-decreasing function, and a late merging strategy (with a higher merging speed $v_M$) will increase the effective discharge rate, which is consistent with our field observation.

2) Scenario 2: reference point $Q$ located on subsegment $\mathcal{MA}$

We now move onto the second scenario where the reference point $Q$ is located on subsegment $\mathcal{DM}$. To differentiate from the previous scenario, we use $Q_2$ to denote this new reference point. As illustrated in **Fig. 4**, since $Q_2$ is located before merging point $\mathcal{A}$, the traffic flow at this location is impacted by the two shockwaves, but no empty space between traffic flows is observed here. In other words, state $TS_2$ does not exist at this location, and the state transition chain is $TS_1 \to TS_3 \to TS_4 \to TS_1$.

Like the analysis of Scenario 1 above, we firstly focus on state $TS_3$, which is the segment between points $V$ and $X$ in Fig. 4, and we use notation $\pi_{Q_2}$ to represent the duration of the orange horizontal line between two shockwaves.

As $\pi_{Q_2}$ represents the duration between points V and X, we have $\pi_{Q_2} = t_X - t_V$. Given a reference point $Q_2$, the time that the queuing shockwave propagates from point $M$ to point $V$ can be computed as $t_{MV} = (x_{Q_2} - x_M)/\omega$, in which $\omega$ is the speed of queuing shockwave, whereas $x_{Q_2}$ and $x_M$ represent the spatial coordinates of these two points. As such, we can obtain that $t_V = t_M + (x_{Q_2} - x_M)/\omega$. On the other hand, the time that the dissipation shockwave propagates from point $A$ to point $X$ can be computed as $t_{AX} = (x_A - x_{Q_2})/w$, in which $w$ is the speed of dissipation shockwave, and $x_A$ and $x_{Q_2}$ represent the spatial coordinates of these two points. As such, we can obtain that $t_X = t_A + (x_A - x_{Q_2})/w$. Combining these two equations from above and with some reorganization, we can have $\pi_{Q_2} = \left(t_A - t_M + \frac{x_A}{w} + \frac{x_M}{\omega}\right) - x_{Q_2}\left(\frac{1}{w} - \frac{1}{\omega}\right)$.

Like the analysis of Scenario 1, during $[t_O, t_V]$, the merging vehicle has no impact on traffic in the freeway mainline, so we have $\mu_{OV} = \mu$, although the actual flow rate might be lower than $\mu$ if the traffic is unsaturated. Next, for the segment $[t_X, t_Y]$, since the ramp vehicle has been fully accelerated, the traffic flow is being discharged at full capacity, so we have $\mu_{XY} = \mu$. For the final segment $[t_Y, t_{O'}]$, the impact of the merging vehicle has disappeared, and traffic on the freeway mainline returns to normal, so we have $\mu_{YO'} = \mu$. The duration of these three periods equals to the episode length $h_r$ minus the duration of segment $VX$, i.e., $(t_V - t_O) + (t_Y - t_X) + (t_{O'} - t_Y) = \frac{1}{\lambda(t)\rho^r} - \left(t_A - t_M + \frac{x_A}{w} + \frac{x_M}{\omega}\right) + x_{Q_2}\left(\frac{1}{w} - \frac{1}{\omega}\right)$.

Finally, by combining these four periods, the effective discharge rate over a complete episode can be computed by $\mu' = (\pi_{Q_2}\mu_{VX} + (t_{OV} + t_{XY} + t_{YO'})\mu)/t_{OO'}$. After plugging in the components from above and some reorganization, we can get a simplified version shown in (10). For a detailed derivation process, please refer to Appendix B.4.

$$\mu' = \mu'_1 + \mu'_2 \tag{10)-(a}$$

$$\mu'_1 = \frac{1}{\lambda(t)\rho^r} - \left(t_A - t_M + \frac{x_A}{w} + \frac{x_M}{\omega}\right)(\mu + \mu_{VX}) \tag{b}$$

$$\mu'_2 = x_{Q_2} \cdot \left(\frac{1}{w} - \frac{1}{\omega}\right) \cdot (\mu - \mu_{VX}) \tag{c}$$

Unlike Scenario 1, the location of reference point $Q_2$, $x_{Q_2}$, remains a variable in (10). In other words, it does not cancel out, indicating that the selection of reference point will impact the derivation of effective discharge rate. According to the 'Barrel Effect' that was discussed at the beginning of this section, the capacity of freeway merging section is determined by its bottleneck, where the capacity is lowest. As such, we need to identify a reference point $Q_2$ where $\mu'$ is the lowest.

We proceed to computing the first-order derivative of (10), and the result is shown below. Interestingly, the first component, $\mu'_1$, is independent of $x_{Q_2}$, so its derivative becomes zero. So, the derivation becomes much easier.

$$\frac{\partial \mu'}{\partial x_{Q_2}} = 0 + \left(\frac{1}{w} - \frac{1}{\omega}\right)(\mu - \mu_{VX}) \tag{11}$$

It is easy to see that $1/w - 1/\omega \leq 0$. This is because $w$ is the speed of the dissipation shockwave, and $\omega$ is the speed of queuing shockwave. As shown in Fig. 4, for traffic flow in a stable state, the slope of dissipation shockwave is steeper than that of queuing shockwave, so we have $w \geq \omega$, and thus, $1/w \leq 1/\omega$. The second item $\mu - \mu_{VX} \geq 0$. This is because as discussed above, $\mu$ is the highest possible discharge rate that is allowed by a fundamental diagram (i.e., the peak of a triangle fundamental diagram), and $\mu_{VX}$ is the discounted flow rate with a higher density.

As such, we have $\partial \mu'_2 / \partial x_{Q_2} \leq 0$, indicating $\mu'$ is a monotonically decreasing function of $x_{Q_2}$. As $x_{Q_2}$ increases (i.e., it's moving away from location $D$ and moving towards point $M$), the effective discharge rate will drop. The lowest discharge rate is observed at point $M$, which has already been derived in (9). The highest discharge rate is observed at point $D$, where $\pi_{Q_2} = 0$ and $\mu' = \mu$. At this location, traffic is not affected by the merging vehicle or any positions further upstream, which is consistent with our field observation.

3) Scenario 3: reference point $Q$ located on subsegment $\mathcal{MA}$

Finally, we move onto the remaining two scenarios. For scenario 3, the state transition chain is $TS_1 \rightarrow TS_1$. It indicates the traffic flow here is unimpacted by the merging vehicle from the on-ramp, as such, we have $\mu' = \mu$. As for scenario 4, its state transition chain is $TS_1 \rightarrow TS_2 \rightarrow TS_4 \rightarrow TS_1$, and it is observed if when reference point $Q$ is placed at a location that is downstream of point $\mathcal{A}$. As illustrated in Fig. 4, at this downstream location, the durations of traffic state $TS_2$ and $TS_3$ both remain a constant. This means the effective discharge rate for this scenario is also a constant and equals to that at location $\mathcal{A}$. In other words, we have $\mu' = \mu'_{\mathcal{A}} = \mu(1 - \lambda(t)\rho^r(t)\frac{(v_u - v_M)^2}{2\alpha v_u})$.

4) Summary of effective discharge rate along the freeway merging area

To summarize, the effective discharge rate along the freeway merging area is dependent on the selection of reference point location, $x_Q$. When this reference point is located before point $\mathcal{D}$, $\mu' = \mu$. When it is located along subsegment $\mathcal{DM}$, $\mu'$ is a monotonically decreasing function of $x_Q$. On the other hand, when this reference point is located along subsegment $\mathcal{MA}$, $x_Q$ cancels out and makes $\mu'$ independent of point $Q$. Finally, after point $\mathcal{A}$, $\mu'$ again remains a constant. As a result, the maximum capacity reduction is observed along subsegment $\mathcal{MA}$, where the ramp vehicle merges into freeway and continue to accelerate to the cruising speed. By using a capacity drop discount, $\theta$, which indicates the degree of discharge rate drop introduced by the merging vehicle, the effective discharge rate along the freeway merging area can be expressed as

$$\mu' = \mu \cdot (1 - \theta) \tag{12}$$

Where $\theta = \lambda(t)\rho^r(t)(v_u - v_M)^2/(2\alpha v_u)$. The capacity discount, $\theta$, is influenced by variables such as arrival flow rate $\lambda(t)$, ramp vehicles ratio $\rho^r(t)$, freeway cruising speed $v_u$, merging speed $v_M$, and acceleration $\alpha$. This highlights the significance of an appropriate merging control strategy when considering its impact on the effective discharge rate.

*B. Formulation of Dynamic Programming Model for Merging Control*

The derivation of the closed form expression for $\mu'$ establishes an analytical relationship between the merging strategy and freeway capacity, which is crucial to the characterization of freeway traffic state and their transitions. Building on this analytical foundation, we formulate freeway merging control as a dynamic programming model to identify the optimal merging strategy and its impact on traffic flow efficiency and driving safety.

Our objective is to jointly minimize traffic flow delay and crash risk resulting from the ramp vehicle's merging decisions. To achieve this, a system cost function, $c_k(s_k)$, is defined that describes the cost associated with the merging vehicle's decision at step $k$ and state $s_k$, as illustrated in (13).

$$c_k(s_k) = m_k \cdot \{\varphi \cdot W_k(s_k) + (1 - \varphi) \cdot \mathcal{S}_k(s_k)\} \tag{13}$$

Where $W_k(s_k)$ and $\mathcal{S}_k(s_k)$ represents the traffic delay and crash risk, respectively, resulting from a vehicle's merging decision $m_k$. Here, $m_k$ is a binary variable, where $m_k = 1$ indicates the vehicle merges into the freeway mainline at step $k$, and $m_k = 0$ indicates it does not merge. The traffic delay is formulated as (2), while the crash risk is computed as (4). The weight parameter, $\varphi$, represents the trade-off between safety and efficiency, where $0 \leq \varphi \leq 1$. When $\varphi = 0$, the formulated dynamic program model optimizes only merging safety, whereas when $\varphi = 1$, minimizing traffic delay becomes the sole objective.

1) Definition of vehicle states, decision variables, and state transition function

We define the state of merging vehicles at step $k$ using a triple $s_k = (l_k, v_k, d_k)$, where: i) $l_k$ denotes the lane index, where $l_k \in I$; ii) $v_k$ represents the merging vehicle's speed, bounded by $v_O \le v_k \le v_u$, where $v_O$ is the initial speed at point $O$ at time $t_O$ (i.e., at point $O$ in the time-space diagram), and $v_u$ is the freeway cruising speed; iii) $d_k$ is the distance from the entrance of the auxiliary lane, defined as $d_k \in D = [0, L + l\,]$, where $l$ is the length of a vehicle. The state space is denoted by $S$, so $s_k \in S$. The initial state of the merging vehicle is given by $s_0 = (l_0, v_O, 0)$, and the terminal state is $s_n = (l_n,\ v_u,\ L + l)$.

The dynamic programming model aims to minimize delays and crash risks associated ramp vehicle merging decisions. To achieve this, two decision variables are defined: $m_k$ for merging decisions and $a_k$ for acceleration decisions. Each vehicle merges only once, expressed as $\sum m_k = 1$. The variable $a_k$ indicates vehicle's acceleration, bounded by $0 \le a_k \le \alpha$ from point $M$ to $A$ in **Fig. 4**, where $\alpha$ is the maximum acceleration based on vehicle performance. After point $A$ (in **Fig. 4**), when the vehicle has fully accelerated, $a_k = 0$.

The state transition function is then defined as a function of the state and decision variables. In (14), the function $T(\cdot)$ represents the state transition from $s_k$ to $s_{k+1}$ when progressing from step $k$ to $k + 1$.

$$s_{k+1} = T(s_k, m_k, a_k) \tag{14}$$

This transition process is further illustrated in **Fig. 6**. As previously mentioned, the initial state of the merging vehicle is denoted as $s_0 = (l_0, v_0, d_0)$ where $l_0 = 1$, $v_0 = v_O$, and $d_0 = 0$ at step $k = 0$. The terminal state is denoted as $s_n = (l_n, v_n, d_n)$ at step $k = n$. Referring to the example illustrated in **Fig. 2**, it is determined that $l_n = 2$, $v_n = v_u$, and $d_n = L + l$. Thus, the terminal state can be expressed as $s_n = (l_n, v_u, L + l)$. Each step $k$ corresponds to an equal time interval of $\Delta t$, and thus the number of steps is given by $n = T/\Delta t$, where $T$ denotes the total duration of the merging process. At each interval $\Delta t$, the arrow in **Fig. 6** always moves downward and points to the succeeding state.

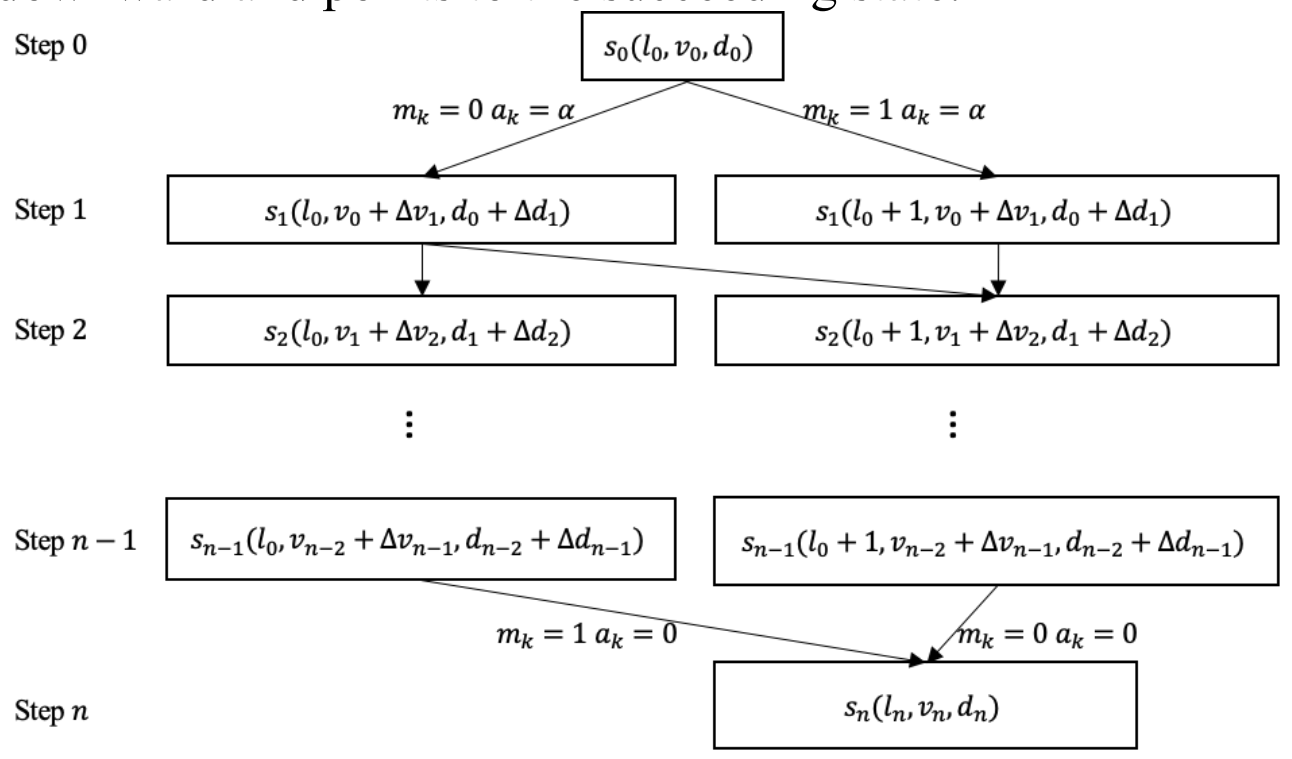

**Fig. 6.** Dynamic state transition of a complete merging process.

Mathematically, the current lane of the vehicle is determined by $l_{k+1} = l_k + m_k$. If the vehicle decides to merge at step $k$ (i.e., $m_k = 1$), the arrow in **Fig. 6** will point to the right-hand block. Leftward transitions are not allowed. For the vehicle's speed, it is updated according to $v_{k+1} = v_k + a_k \Delta t$, where the acceleration is constrained by $0 \le a_k \le \alpha$ during the interval $t_M \le k\Delta t \le t_A$, and $a_k = 0$ afterward. Regarding the vehicle's position, as illustrated in Fig. 6, it is given by $d_{k+1} = d_k + \Delta d_k$, where $\Delta d_k = v_k \Delta t + \frac{a_k}{2} \Delta t^2$.

2) State transition-based dynamic programming model for merging control

This subsection presents the dynamic programming formulation for optimized decision-making and discusses the constraints on two decision variables. The complete modeling formulation is then summarized.

In the terminal state, the ramp vehicle will eventually switch lanes to the freeway mainline and fully accelerate to match the prevailing freeway speed. The ramp vehicle merging task is formulated as a recursive optimization problem, solved using backward induction to find the minimum-cost solution. This method begins at the final step of the merging sequence and works backward, optimizing decisions at each stage to minimize overall costs. By considering future impacts at each step, this approach ensures a globally optimized solution for the entire merging process.

The recursive optimization problem on this minimum-cost problem is formulated as:

$$f(s_k) = min\big(c_k(s_k) + f(s_{k+1})\big) \tag{15}$$

The terminal state, $s_n$, as illustrated in **Fig. 6**, is located at the bottom of the state transition diagram and indicates the merging vehicle has been fully accelerated on the freeway mainline. As such, the cost associated with that final state is 0, as shown in (16).

$$f(s_n) = 0 \tag{16}$$

For the merging decision $m_k$, a ramp vehicle merges into the freeway mainline exactly once. Since $m_k$ is a binary variable, it satisfies $\sum m_k = 1$. When an acceptable gap is observed on the freeway mainline, the merging vehicle has two options: initiate a lane change (represented by an arrow pointing to the right in **Fig. 6**, with $m_k = 1$) or reject the gap and continue accelerating on the auxiliary lane (representing by an arrow pointing downward, with $m_k = 0$ ). The probability of finding an acceptable gap for merging, $P_M(t)$, can be derived by (1). In other words, the probability of not finding an acceptable gap is $1 - P_M(t)$, in which case the vehicle is not allowed to merge. When an acceptable gap appears with a probability of $P_M(t)$, the value of $m_k$ (either 0 or 1) is determined based on the recommended action from the optimization model, aiming to minimize traffic flow impact. It is important to note that if a ramp vehicle reaches the end of auxiliary lane without finding a satisfactory gap, a mandatory lane change will be initiated regardless of the headway gap. However, such a maneuver raises significant safety concerns. The impact of these decisions is illustrated in the dynamic programming model presented later. Regarding acceleration decisions $a_k$, after reaching point $M$, the merging vehicle begins to accelerate at a rate within the range $(0, \alpha)$ until it is fully accelerated at point $A$. After this point, $a_k = 0$, and the vehicle maintains a speed consistent with the prevailing freeway traffic flow conditions.

Summarizing the analysis above, the formal formulation of this dynamic programming problem is shown below.

$$f(s_k) = min\big(c_k(s_k) + f(s_{k+1})\big) \tag{17-(a)}$$

s.t.

$$f(s_n) = 0 \tag{b}$$

$$c_k(s_k) = m_k \cdot \{\varphi \cdot W_k(s_k) + (1-\varphi) \cdot \mathcal{S}_k(s_k)\} \tag{c}$$

$$s_{k+1} = T(s_k, m_k, a_k) \tag{d}$$

$$l_{k+1} = l_k + m_k, v_{k+1} = v_k + a_k \cdot \Delta t, d_{k+1} = d_k + \Delta d_k \tag{e}$$

$$\Delta d_k = v_k \Delta t + \frac{a_k \Delta t}{2}, \begin{cases} 0 \le a_k \le \alpha, if\ t_M \le k\Delta t \le t_A \\ a_k = 0 \qquad\qquad otherwise \end{cases} \tag{f}$$

$$P_M(k) = e^{-\left(\lambda \cdot \frac{\rho^m}{3600}\right)\cdot\left(\tau + \frac{v_u}{\alpha} + \frac{v_k}{\alpha}\right)} \tag{g}$$

$$\mu' = \mu(1 - \lambda(t)\rho^r(t)\frac{(v_u - v_M)^2}{2\alpha v_u}) \tag{h}$$

$$Q'(t) = \int_{t_O}^{t} (\lambda(\tau) - \mu')d\tau \tag{i}$$

$$W(\mu') = \int_{t_O}^{t_{O'}} Q'(t)d\tau \tag{j}$$

$$\mathcal{S} = \int_{t_O}^{t_{O'}} \left( \frac{\big(v_{i+1}(t) - v_j(t)\big)^2}{x_j(t) - x_{i+1}(t) - L_j} + \sum_{k=1}^{K} \frac{\big(v_{i+k+1}(t) - v_{i+k}(t)\big)^2}{x_{i+k}(t) - x_{i+k+1}(t) - L_{i+k}} \right) dt \tag{k}$$

$$m_k \in \{0,1\}, \sum m_k = 1, l_k \in \{1,2\} \tag{l}$$
$$v_k \in [v_0, v_{\mathrm{u}}], d_k \in [0, L + l\,]$$

In (17), (a) is the recursive Bellman equation to optimize the AV's merging decisions at each step. Constraint (b) indicates the terminal state, while constraint (c) describes the weighted cost at each state. (d)~(f) represent state transition functions, and (g) is the probability of finding an acceptable gap. (h)~(j) are equations to compute traffic flow efficiency, and (k) serves as the surrogate traffic safety measurement. (l) defines the possible range for each variable. Given the specified terminal state in this problem, the backward induction method is best suited to solve this problem.

## V. Numerical Analysis

This section presents numerical analysis to validate the proposed model and assess its performance. The study is guided by a key principle: to first develop a foundational physical model applicable to both AVs and HDVs, validate it using the NGSIM dataset and then leverage the validated model to optimize AV decision-making.

### *A. Validation of Effective Discharge Rate*

To validate the closed-form formulation of the effective discharge rate, we utilize two NGSIM datasets collected from Northbound I-80 in California on April 13, 2005. The study area spans approximately 500 m and includes a 43 m auxiliary lane with an on-ramp on the freeway's right side as shown in **Fig. 7**. The analysis centers on the impact of merging vehicles on the effective discharge rate of the freeway's mainline, thus, two lanes are selected for evaluation: lane 6 on the freeway and lane 7 on the ramp.

An open-source software DTALite/NeXTA [45] is used to process NGSIM vehicle trajectories and extract variables such as jam density and backward wave speed. A time-space diagram in **Fig. 8** shows vehicle trajectories from 4:00 to 4:15 p.m., highlighting shockwave propagation in areas A, B, and C. The slopes of these shockwaves range from 16 km/hr to 22 km/hr, with an average backward wave speed of 19 km/hr. The diagram also reveals a jam density of 113 vehicles/lane/km and a cruising speed of 48 km/hr.

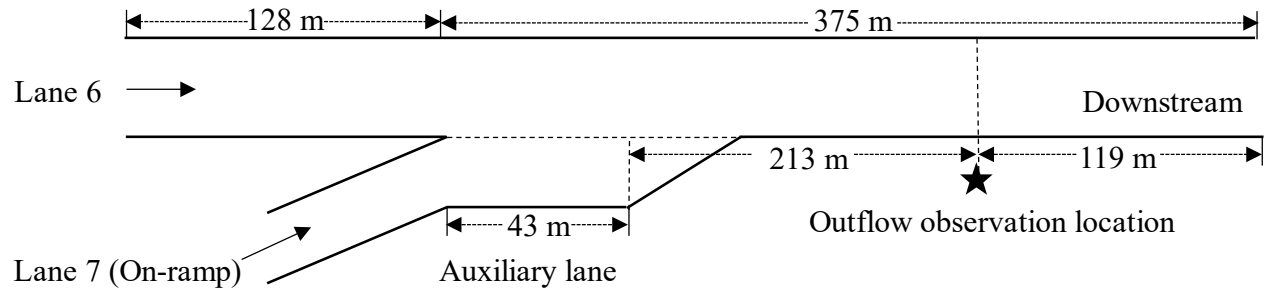

**Fig. 7.** An illustration of the study area, Northbound I-80, CA.

The maximum discharge rate is calculated based on the triangle fundamental diagram as $\mu = 1{,}538$ veh/h. The freeway mainline arrival rate is measured upstream of the merging area. The combined vehicle arriv18al rate on lanes 6 and 7 is $(265 + 192) \times 4 = 1{,}828$ veh/h. On-ramp vehicles accelerate along a 43 m-long auxiliary lane while seeking opportunities to merge. An additional 213 m is included to establish an observation point downstream, ensuring vehicles are fully accelerated before reaching this location.

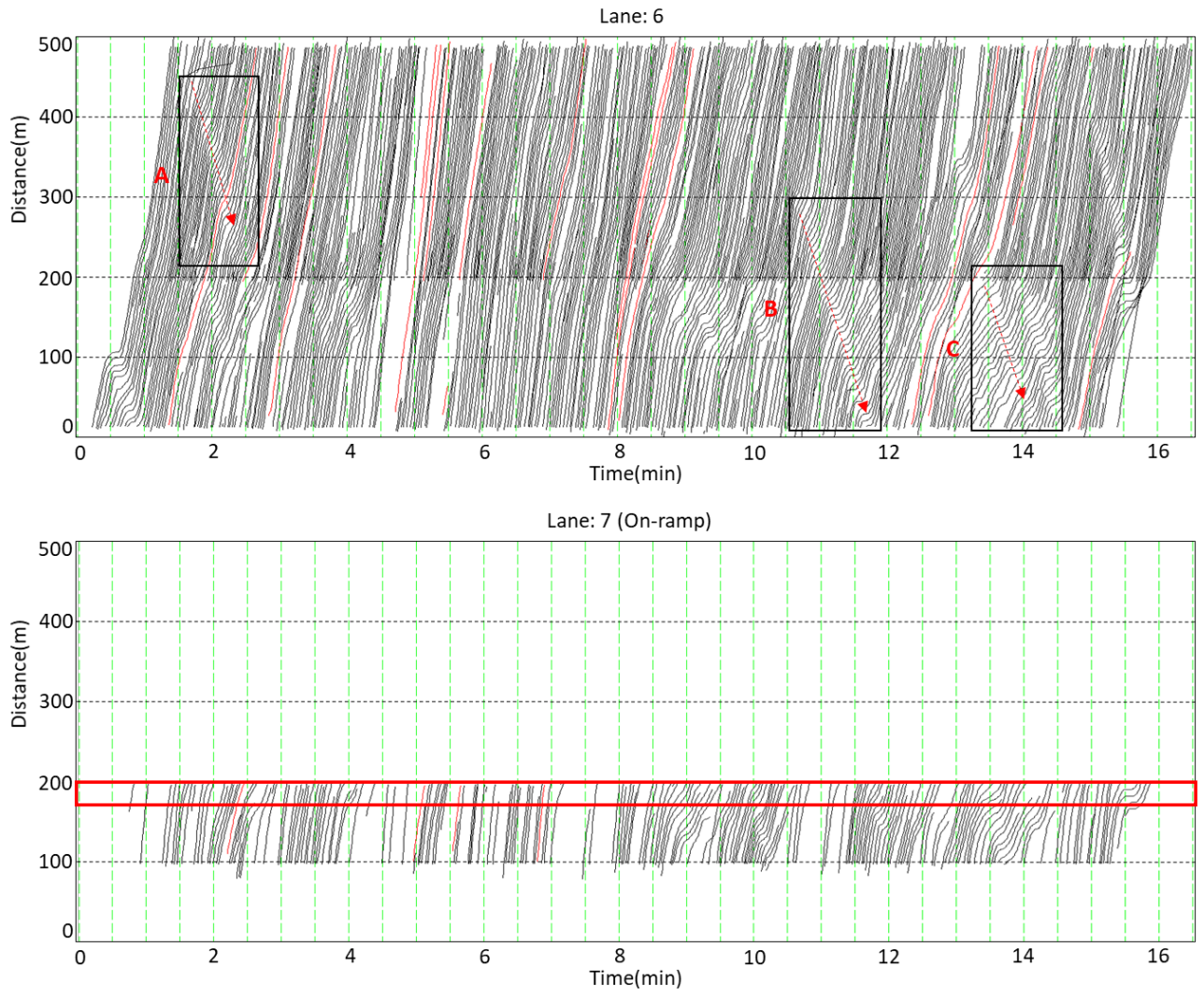

**Fig. 8.** Vehicle trajectories from NGSIM dataset for lanes 6 and 7.

The NGSIM dataset shows that 280 vehicles pass the downstream point on lane 6 within 15 minutes, equating to a ground truth discharge rate of 1,120 veh/hr. Examination of the NGSIM on-ramp merging vehicle trajectory an average acceleration of approximately $1.5\ \mathrm{m/s^2}$, and the average merging speed on lane 7 is estimated to be 24 km/hr. The effective discharge rate, $\mu'$, is calculated as $\mu' = 1538 \times (1 - 0.233) = 1{,}174$ veh/hr. The absolute percentage error (APE) of the derived effective discharge rate compared to the ground truth discharge rate is computed as $\mathrm{APE}_{\mu'} = 4.8\%$. In contrast, if the maximum discharge rate $\mu$ was mistakenly used, the APE would be $\mathrm{APE}_{\mu} = 37.3\%$, which is significantly higher and could lead to problematic decision-making.

A second validation is performed using the NGSIM dataset from 5:00 to 5:30 p.m. on I-80, following a similar approach to the first. The combined arrival rate for the freeway mainline and on-ramp is $(461 + 399) \times 2 = 1{,}720$ veh/hr. The observed ground truth discharge rate is $(371 + 276) \times 2 = 1{,}294$ veh/hr. The average merging speed is observed to be 26 km/hr, and the effective discharge rate, $\mu'$, is determined to be $\mu' = 1538 \times (1 - 0.20) = 1{,}230$ veh/hr. Compared to the ground truth data, the APE of the derived effective discharge rate, $\mu'$, can be computed as $APE_{\mu'} = 4.9\%$. In contrast, if the maximum discharge rate, $\mu$, is mistakenly used, the APE would be $APE_{\mu} = 18.9\%$, which is much higher and will lead to problematic decision-making.

*B. Performance Evaluation*

1) Case study setup

After validating the closed-form effective discharge rate formulation, the performance of the proposed AV merging control model is assessed. For numerical analysis, a simple network is designed, comprising a freeway and a single-lane on-ramp merging into the freeway from the right. The total length of the study area is 600 m, with the auxiliary lane spanning 150 m. The study period lasts 150s. The cruising speed on the freeway mainline, $v_u$, is set to be $105$ km/hr and the backward wave speed is $w = 16$ km/hr. Additionally, the speed limit of the on-ramp is set at $v^r_{max} = 56$km/hr, the maximum acceleration is set at $2\ \mathrm{m/s^2}$, and the maximum deceleration rate is set at $-6\mathrm{m/s^2}$. The reaction time for braking is set at 1.5 s. Vehicle demand is set at 1,600 veh/hr, with 15% of the vehicles entering from the on-ramp.

2) Performance comparison with and without calibrated effective discharge rate

The dynamic programming model in (17) was solved using Python. Due to the stochasticity in the traffic flow, 1,000 experiments were conducted with randomly generated headway gaps between consecutive freeway vehicles, solving the optimization problem for each instance. Traffic delay and crash risk values were normalized to a 0–1 range for comparison.

This subsection compares results using the calibrated effective discharge rate, $\mu'$, with those obtained by mistakenly using the maximum discharge rate $\mu$. In the latter case, the same model as in (17) was used, but $\mu$ replaced $\mu'$ in (i)-(j), and constraint (h) was omitted. TABLE 1 summarizes the comparison results. The crash risk index $S$ remains almost unchanged regardless of whether $\mu'$ is utilized. However, when ignoring the effective discharge rate and mistakenly applying the maximum discharge rate, the traffic delay $W$ is significantly underestimated. Specifically, with an efficiency-oriented objective ($\varphi = 1$), the weighted cost is underestimated by 4.4%. When efficiency and safety are given equal importance ($\varphi = 0.5$), the underestimation is 3.3%.

TABLE 1

MODELING RESULTS COMPARISON WITH AND WITHOUT CALIBRATED EFFECTIVE DISCHARGE RATE

| | Modeling results with effective discharge rate $\boldsymbol{\mu}'$ | | | Modeling results with maximum discharge rate $\boldsymbol{\mu}$ | | |
|---|---|---|---|---|---|---|
| | Delay (veh-s) | Crash risk | Weighted cost $\boldsymbol{c}$ | Delay (veh-s) | Crash risk | Weighted cost $\boldsymbol{c}$ |
| Efficiency oriented ($\boldsymbol{\varphi = 1}$) | 120 | 5.37 | 0.3466 | 115 (4.4%) | 5.37 (-) | 0.3319 (3.3%) |
| Safety oriented ($\boldsymbol{\varphi = 0}$) | 122 | 5.33 | 0.1160 | 117 (4.3%) | 5.33 (-) | 0.1160 (-) |
| Both considered ($\boldsymbol{\varphi = 0.5}$) | 120 | 5.35 | 0.2316 | 115 (4.4%) | 5.36 (-) | 0.2242 (4.4%) |

3) Performance comparison with two benchmark models

Additionally, the benefits of using $\mu'$ in the merging decision-making were analyzed in comparison to two rule-based models. These benchmark models are based on two strategies: "*early-merging strategy*" where merging vehicles merge into the freeway once an acceptable gap is found [4]; and "*late-merging strategy"* where merging vehicles merge only at the end of the auxiliary lane [5]. Since the source code of these two algorithms was not available, efforts were made to reproduce their methods to the maximum degree possible. It should be noted that these benchmark models were coded based on the key ideas and steps that are described in the papers, and due to the lack of source code, they may not be entirely identical to the actual models. Some simplifications were also made, for instance, in this study the AVs are assumed to make their merging decisions independently, so the cooperative decision-making was not examined.

Three scenarios were analyzed: an efficiency-oriented scenario ($\varphi = 1$, focused on reducing traffic delay), a safety-oriented scenario ($\varphi = 0$, prioritizing crash risk reduction), and a third scenario ($\varphi = 0.5$, equally considering delay and crash risks). The results, normalized as weighted cost $c$, are presented, with the average cost reduction compared to benchmark models shown in brackets.

TABLE 2

MODELING RESULTS COMPARISON WITH TWO BENCHMARK MERGING STRATEGIES

| | **Proposed model** | | | **Early-merging strategy** (Letter and Elefteriadou 2017) | | | **Late-merging strategy** (Ntousakis et al. 2016) | | |
|---|---|---|---|---|---|---|---|---|---|
| | Delay (veh-s) | Crash risk | Weighted cost $c$ | Delay (veh-s) | Crash risk | Weighted cost $c$ | Delay (veh-s) | Crash risk | Weighted cost $c$ |
| Efficiency oriented ($\varphi = 1$) | 120 | 5.37 | 0.3466 | 141 (+15.3%) | 6.62 (+19.1%) | 0.4090 (+15.3%) | 131 (+8.6%) | 5.77 (+6.9%) | 0.3793 (+8.6%) |
| Safety oriented ($\varphi = 0$) | 122 | 5.33 | 0.1160 | 141 (+13.7%) | 6.62 (+19.7%) | 0.1444 (+19.7%) | 131 (+7.0%) | 5.77 (+7.6%) | 0.1255 (+7.6%) |
| Both considered ($\varphi = 0.5$) | 120 | 5.35 | 0.2316 | 141 (+15.2%) | 6.62 (+19.3%) | 0.2767 (+16.3%) | 131 (+8.6%) | 5.77 (+7.2%) | 0.2524 (+8.2%) |

The results clearly show that the proposed model outperforms both benchmark models in all aspects. Specifically, when compared to the early-merging strategy [4], the proposed model achieves an average reduction in delay and crash risks of 15.3% and 19.1% respectively, under an efficiency-oriented goal. With a safety-oriented goal, the proposed model achieves reductions of 13.7% and 19.7%. Even when both objectives are given equal importance, the proposed model continues to outperform, achieving reductions of 15.2% and 19.3%. Additionally, the weighted cost, $c$, reduces by 15.3%, 19.7%, and 16.3% under these three different goals.

Compared to the late-merging strategy [5], the proposed model also demonstrates superior performance. With an efficiency-oriented goal, the average reductions in delay and crash risks stand at 8.6% and 6.9%, respectively. Conversely, for a safety-oriented goal, these reductions are 7.0% and 7.6%. When both objectives are given equal weight, the proposed model achieves reductions of 8.6% and 7.2%, respectively. Moreover, the weighted cost, $c$, reduces by 8.6%, 7.6%, and 8.2% respectively. Based on these findings, it can be inferred that the optimized merging decisions obtained from the proposed model consistently result in superior traffic flow performance in terms of both efficiency and safety, as compared to the two benchmark models.

*C. Sensitivity Analysis*

This section presents a sensitivity analysis to investigate how input data affects the proposed model's performance. Specifically, we examine the impact of the demand, the length of the auxiliary lane, and the demand ratio of on-ramp traffic. The outcomes of these tests are summarized below.

1) When demand changes

The demand, $\lambda(t)$, varied from 1,200 veh/hr to 2,200 veh/hr, in increments of 100. Corresponding changes in the performance results are depicted in **Fig. 9**. The X-axis denotes the demand, while the Y-axis represents the percentage of average cost reduction in the performance measurements. The three red lines show the performance outcomes compared to the early-merging strategy, and the three blue lines show the performance results compared to the late-merging strategy. The solid, dashed, and dotted lines indicate the performance results for traffic efficiency-oriented ($\varphi = 1$), safety-oriented ($\varphi = 0$), and a combination of both efficiency and safety ($\varphi = 0.5$), respectively. As can be observed from the figure, irrespective of the demand fluctuations, the proposed model consistently outperforms the two benchmark models, as indicated by the positive reduction in the performance measurements. The proposed model can reduce costs by 0~22.1% and 0.5%~15.2% compared to the early-merging strategy and late-merging strategy, respectively.

Compared to the late-merging strategy, the gain becomes less significant as the demand increases. This could be due to the increasing flow rate, which likely reduces the probability of finding an acceptable gap. Consequently, vehicles might accelerate initially on the auxiliary lane and merge later. With limited merging opportunities available, the proposed model is more likely to perform merging at the last minute. Another notable observation is that when demand is lower than 1400 veh/hr and the efficiency-oriented goal is adopted, the proposed model achieves the same outcomes as the early-merging strategy. This may be because when the demand is low, it is relatively easy to find a larger gap that enables the ramp vehicle to merge into the freeway while not impacting those driving on the freeway mainline. Consequently, the proposed model also leads to an early-merge strategy, which is essentially the same as the early-merging strategy. However, as demand increases, the likelihood of finding a suitable gap decreases and merging without full acceleration discounts the capacity. In such context, the early-merging strategy leads to noticeable delays for other vehicles. Consequently, the average cost reduction sees a sudden increase as demand rises from 1400 veh/hr to 1600 veh/hr.

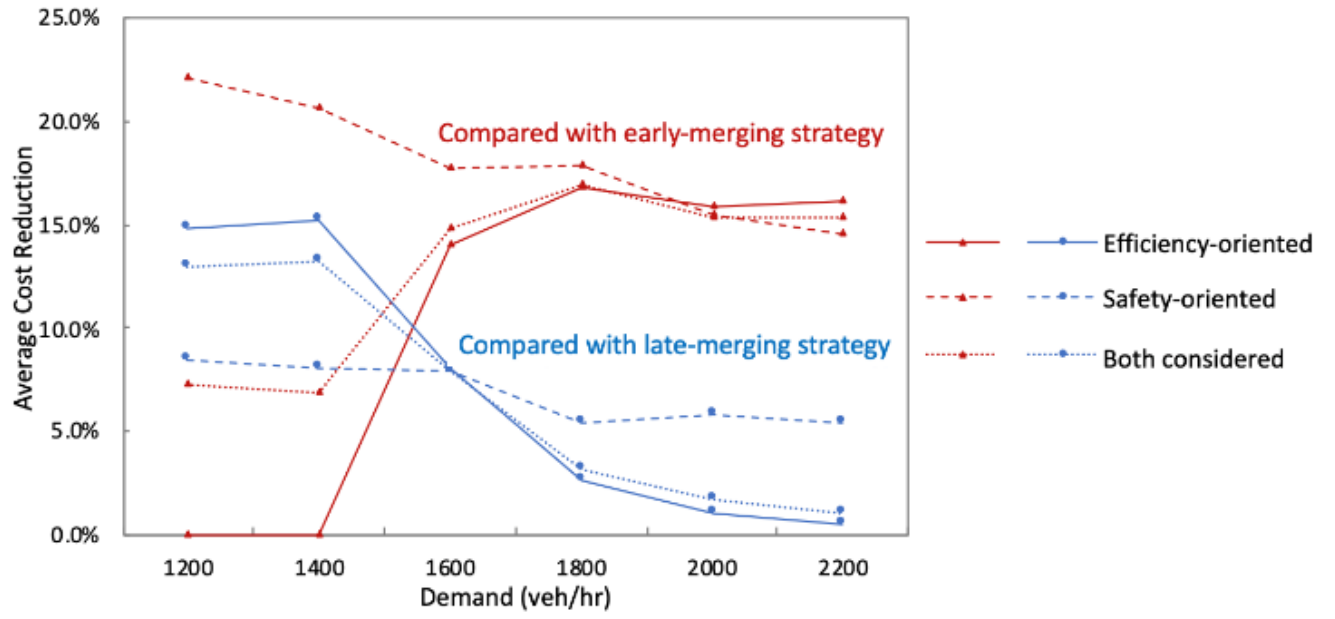

**Fig. 9.** Performance results compared with early-merging strategy and late-merging strategy when demand changes.

2) When the length of the auxiliary lane changes

In the next step, the auxiliary lane length was increased from 100 m to 200 m in 10 m increments. **Fig. 10** shows the resulting performance change. Similar to the previous analysis, six curves – three for each benchmark strategy – are plotted for comparison. The results show that the proposed model consistently outperforms both benchmark models, with reductions in performance costs observed across all auxiliary lane lengths. The proposed model achieves cost reduction ranging from 12.1% to 56.3% compared to the early-merging strategy and 0% to 10.7% compared to the late-merging strategy. As the auxiliary lane length increases, the efficiency-oriented curves remain overall stable, with a slight decrease.

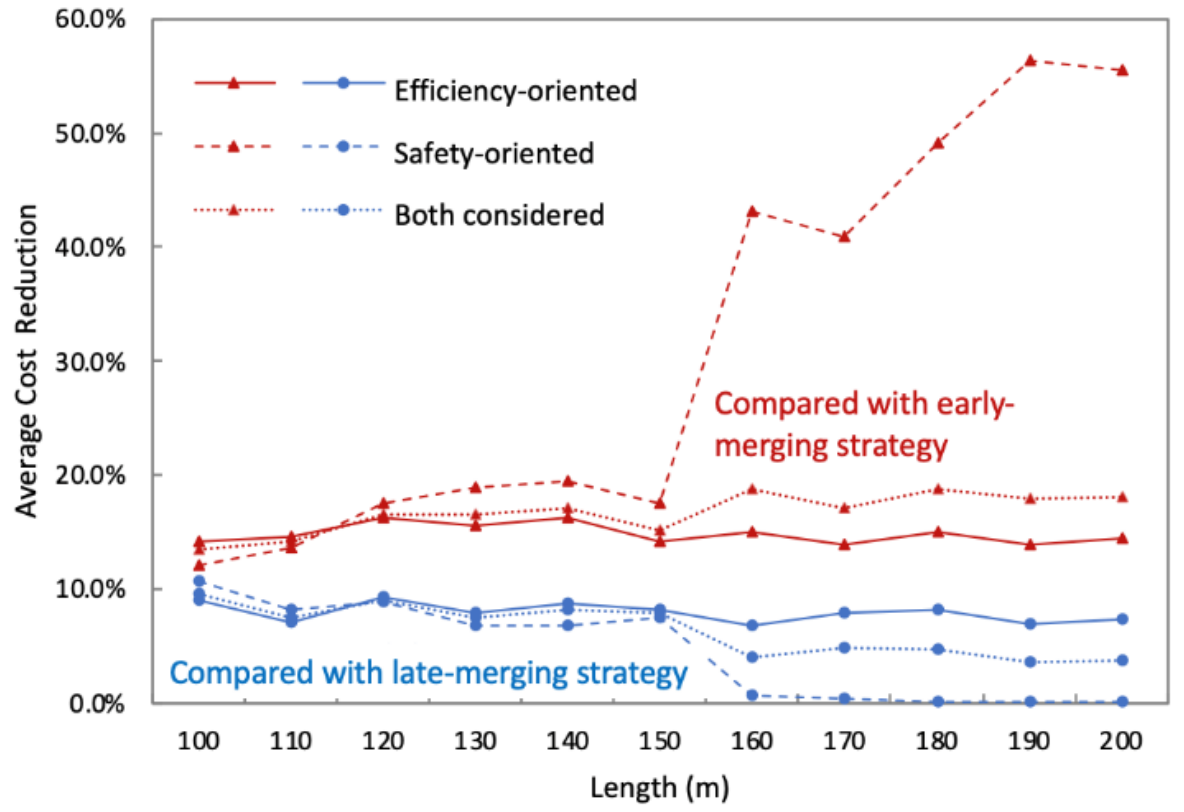

**Fig. 10.** Performance results when the length of the auxiliary lane changes.

Furthermore, the two efficiency-oriented curves remain stable at 14.8% and 7.9%. However, the safety-oriented curve shows a significant increase compared to the early-merging strategy, particularly when the auxiliary lane exceeds 150 m. This is likely because vehicles have more opportunities to find acceptable gaps after reaching cruising speed, enabling safer merges. Moreover, little to no safety improvement is observed compared to the late-merging strategy, where vehicles merge at cruising speed.

3) When the demand ratio of the on-ramp changes

Next, we adjust the ratio of vehicles from on-ramps ($\rho^r$), while keeping the total demand constant. The percentage of ramp vehicles varies from 5% to 20% in 5% increment. **Fig. 11** shows the changes in performance results under these adjustments, with legends consistent with previous analyses.

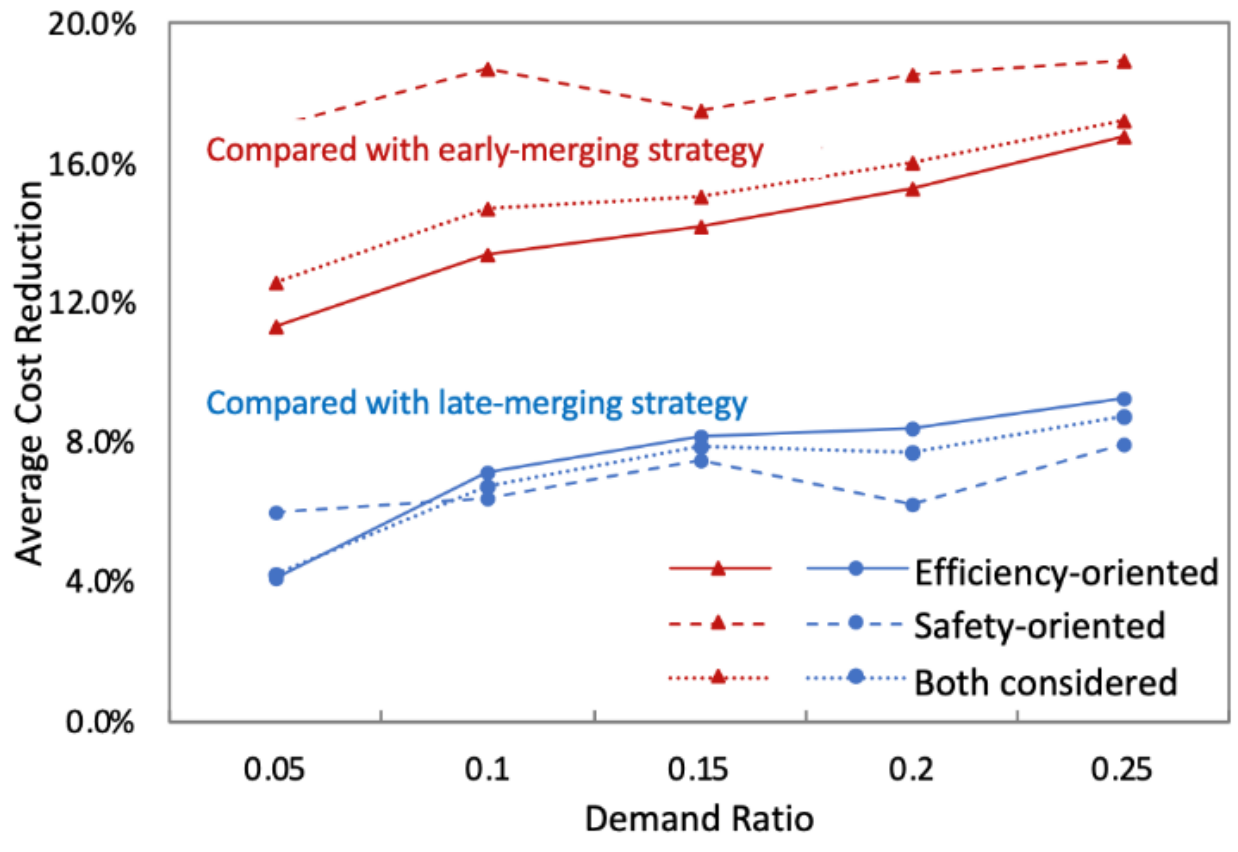

**Fig. 11.** Performance results when the demand ratio of on-ramp changes.

The proposed model consistently outperforms the two benchmark models, achieving cost reductions of 11.3~18.9% compared to the early-merging strategy and 4.1~9.3% compared to the late-merging strategy. As the ratio of ramp vehicles increases, all six curves show a slight upward trend, highlighting that the proposed model's benefits become more pronounced with higher merging vehicle ratios. This is likely because a higher number of merging vehicles leads to an increase in rational agents with optimized trajectories, enhancing the overall performance of the proposed model.

## VI. Conclusion

This manuscript proposes an analytical approach to optimize the dynamic multi-stage merging process of autonomous vehicles, balancing traffic flow efficiency and safety. The dynamic merging process is characterized, including the selection of merging points, the state of merging vehicles, and their transitions. The core modeling efforts focus on deriving an effective discharge rate at merging locations using a closed-form expression, followed by the derivation of traffic delay and crash risks. A dynamic programming problem is then formulated to optimize the autonomous vehicle's merging decisions, tightly integrating traffic flow efficiency and safety objectives.

Numerical experiments using NGSIM dataset were conducted to validate the model. The results show that the proposed model achieves an effective discharge rate calibration with accuracies of 4.8% and 4. when compared to two field datasets. In contrast, using the maximum discharge rate mistakenly would result in errors of 37.3% and 18.9%. Further analysis demonstrates the proposed model's capability to improve traffic flow efficiency and safety, achieving system cost reductions of 13.7% to 19.7% and 6.9% to 8.6%, respectively, compared to the early-merging strategy and the late-merging strategy. Lastly, a sensitivity analysis assessed the impact of input data, such as demand, auxiliary lane length, and on-ramp traffic ratio, on the model's performance. The results consistently demonstrate that the proposed model outperforms both benchmark models in all aspects.

A limitation of this research is that focusing on a single automated vehicle and a single lane does not fully capture real-world scenarios, potentially hindering practical applications. Further research will expand to include collaboration among connected and automated vehicles, multi-lane freeway scenarios, and the incorporation of lane-changing maneuvers for greater realism. This expansion will introduce additional complexity, significantly enlarging the state space and presenting new challenges for solving recursive optimization in a multi-lane context.

## Appendices

### *Appendix A – Notation List*

| | |
|---|---|
| $g_{lag}, g_{lead}$ | Lag and lead gap distance |
| $v_0, v_M$ | Speed of a merging vehicle arrives at the entrance of the auxiliary lane, and merging speed |
| $v_u$ | Cruising speed on the mainline |
| $\lambda(t)$ | Overall arrival rate |
| $\rho^r, \rho^m$ | Ratio of vehicles from the on-ramp and freeway vehicles |
| $\alpha, \tau$ | Vehicle's maximum deceleration and driver's response time |
| $t^c_{\text{lag/lead}}(t)$ | Required lag or lead gap for the merging at time $t$ |
| $h_M(t)$ | The required time headway for merging at time $t$ |
| $P_M(t)$ | Probability of finding an acceptable gap for at time $t$ |
| $s_k$ | State of the merging vehicle at step $k$ |
| $d_k, l_k, v_k$ | Distance of the merging vehicle from the entrance of auxiliary lane, lane index, and merging vehicle's speed |
| $m_k$ | Merging decision variable |
| $w, k_j$ | Speed of the backward wave and Jam density |
| $\mu, \mu'$ | Maximum discharge rate and effective discharge rate |
| $Q(t), Q'(t)$ | Queue length at time $t$ using $\mu$ and $\mu'$ |
| $W(\mu), W(\mu')$ | Total delay if ignoring or considering capacity drop |
| $S$ | Crash risk function |
| $c_k(s_k)$ | Cost associated with the merging vehicle's decision at step $k$ |
| $\varphi$ | A weight parameter, $0 \le \varphi \le 1$ |

### *Appendix B – Derivation*

*B.1 Derivation of (7)*

$$t_S - t_A = \frac{x_{SA}}{w} = \frac{x_{Q_1A}}{w} = \frac{v_M \cdot t_{Q_1A} + \frac{1}{2}\cdot\alpha\cdot {t_{Q_1A}}^2}{w}$$

$$= \frac{v_M \cdot \left(\frac{v_u - v_M}{\alpha}\right) + \frac{1}{2}\cdot\alpha\cdot\left(\frac{v_u - v_M}{\alpha}\right)^2}{w} = \frac{v_u^2 - v_{Q_1}^2}{2\alpha w}$$

*B.2 Derivation of (8)*

$$\mu_{Q_1S} = wk_j \frac{t_S - t_A}{\pi_{Q_1}} = wk_j \frac{\frac{v_u^2 - v_{Q_1}^2}{2\alpha w}}{\frac{v_u^2 - v_{Q_1}^2}{2\alpha w} + \frac{v_u - v_{Q_1}}{\alpha}}$$

*<u>B.3 Derivation of (9)</u>*

$$\mu' = \frac{N_{OO'}}{t_{OO'}} = \frac{t_{OR}\mu_{OR} + \sigma_{Q_1}\mu_{RQ_1} + \pi_{Q_1}\mu_{Q_1S} + t_{SU}\mu_{SU} + t_{UO'}\mu_{UO'}}{t_{OO'}}$$

$$= \left(\sigma_{Q_1}\mu_{RQ_1} + \pi_{Q_1}\mu_{Q_1S} + (t_{OR} + t_{SU} + t_{UO'})\mu\right)/t_{OO'}$$

$$= \Big(\mu\big((t_R - t_O) + (t_U - t_S) + (t_{O'} - t_U)\big) + 0\sigma_{Q_1} + \mu_{Q_1S}\pi_{Q_1}\Big)/h_r$$

$$= \frac{\mu}{h_r}\left(h_r - \frac{v_u - v_M}{\alpha} + \frac{v_{Q_1}^2 - v_M^2}{2\alpha v_u} - \frac{v_u^2 - v_{Q_1}^2}{2\alpha w} + \frac{v_u + w}{k_j v_u w} k_j \frac{v_u^2 - v_{Q_1}^2}{2\alpha}\right)$$

$$= \frac{\mu}{h_r}\left(h_r - \frac{v_u - v_M}{\alpha} + \frac{v_u^2 - v_M^2}{2\alpha v_u}\right) = \mu\left[1 - \lambda(t)\rho^r(t)\frac{(v_u - v_M)^2}{2\alpha v_u}\right]$$

*B.4 Derivation of (10)*

$$\mu' = \frac{(t_{OV} + t_{XY} + t_{YO'})\mu + \pi_{Q_2}\mu_{VX}}{t_{OO'}}$$

$$= \left(\frac{1}{\lambda(t)\rho^r} - \left(t_{\mathrm{A}} - t_{\mathrm{M}} + \frac{x_{\mathrm{A}}}{w} + \frac{x_{\mathrm{M}}}{\omega}\right) + x_{Q_2}\left(\frac{1}{w} - \frac{1}{\omega}\right)\right)\mu$$

$$+ \left(\left(t_{\mathrm{A}} - t_{\mathrm{M}} + \frac{x_{\mathrm{A}}}{w} + \frac{x_{\mathrm{M}}}{\omega}\right) - x_{Q_2}\left(\frac{1}{w} - \frac{1}{\omega}\right)\right)\mu_{VX}$$

$$= \left(\frac{1}{\lambda(t)\rho^r} - \left(t_{\mathrm{A}} - t_{\mathrm{M}} + \frac{x_{\mathrm{A}}}{w} + \frac{x_{\mathrm{M}}}{\omega}\right)\mu + \left(t_{\mathrm{A}} - t_{\mathrm{M}} + \frac{x_{\mathrm{A}}}{w} + \frac{x_{\mathrm{M}}}{\omega}\right)\mu_{VX}\right)$$

$$+ x_{Q_2}\left(\frac{1}{w} - \frac{1}{\omega}\right)(\mu - \mu_{VX})$$

If we let ${\mu'}_1 = \frac{1}{\lambda(t)\rho^r} - \left(t_{\mathrm{A}} - t_{\mathrm{M}} + \frac{x_{\mathrm{A}}}{w} + + \frac{x_{\mathrm{M}}}{\omega}\right)\mu + \left(t_{\mathrm{A}} - t_{\mathrm{M}} + \frac{x_{\mathrm{A}}}{w} + \frac{x_{\mathrm{M}}}{\omega}\right)\mu_{VX}$, and ${\mu'}_2 = x_{Q_2}\left(\frac{1}{w} - \frac{1}{\omega}\right)(\mu - \mu_{VX})$. Then we will have $\mu' = {\mu'}_1 + {\mu'}_2$.

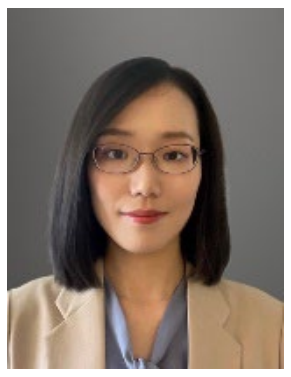

**Qing Tang** is currently an Assistant Professor with the Old Dominion University, in its Civil and Environmental Engineering Department. She earned her PhD in Civil Engineering from Pennsylvania State University, University Park, PA in 2023. Her research focuses on connected and automated vehicles, human-autonomous vehicle interaction, traffic flow modeling, traffic simulation, and optimization algorithms.

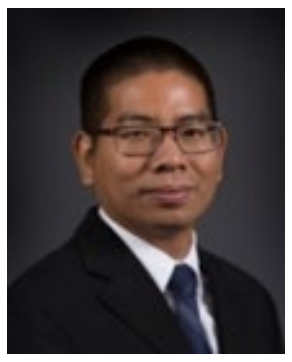

**Xianbiao Hu** is currently an Assistant Professor with the Pennsylvania State University, in its Civil and Environmental Engineering Department. He received the Ph. D. degree from the University of Arizona, Tucson, AZ in 2013. His current research focuses on smart mobility systems, connected and automated vehicles, electric vehicles, mobility behavior management, transportation big data analytics, and traffic flow and system modeling. He is the Committee Research Coordinator of the TRB Committee on Maintenance and Operations Management (AKR10). Additionally, he is an Associate Editor for IEEE Transactions on Intelligent Transportation Systems, an Assistant Editor for the Journal of Intelligent Transportation Systems, and a handling editor for Transportation Research Record.